\documentclass[a4paper,fleqn]{cas-dc}

\usepackage[numbers,sort&compress]{natbib}
\usepackage{booktabs}

\usepackage{threeparttable}
\usepackage[utf8]{inputenc}
\usepackage{newunicodechar}
\usepackage{threeparttable}
\usepackage{tabularx}
\usepackage{subcaption}
\usepackage{placeins}
\usepackage{placeins}
\newunicodechar{；}{;}
\newunicodechar{）}{)}
\usepackage{graphicx} 
\usepackage{subcaption}
\usepackage{amssymb}
\usepackage{amsmath,amssymb,amsfonts}
\usepackage{graphicx}
\usepackage{algorithm}
\usepackage{algorithmic}
\usepackage{amsmath}
\usepackage{multirow}
\usepackage{textcomp}
\usepackage{xcolor}
\usepackage{amsmath}
\usepackage{soul} % 导入 soul 包
\usepackage{color, xcolor}
\usepackage[figuresright]{rotating}
\usepackage[mathscr]{eucal}
\usepackage{bm} 

\usepackage{times}
\usepackage{url}
\usepackage[utf8]{inputenc}
\usepackage[font=small]{caption} % 修复 caption 包的 small 选项问题
\usepackage{amsthm}
\usepackage{booktabs}
\usepackage[switch]{lineno}
\usepackage{adjustbox}
\usepackage{mdframed}
\sethlcolor{yellow} 
\soulregister\cite7 
\soulregister\ref7
\soulregister{\mathbf}{1} % 注册 \mathbf
\soulregister{(}{1} % 注册 \(
\soulregister{)}{1} % 注册 \)
\newmdenv[
    backgroundcolor=yellow!100, % 背景色
    linecolor=yellow!100, % 边框色
    skipabove=1em, % 上方间距
    skipbelow=1em, % 下方间距
    nobreak=true % 支持跨页
]{highlighted}

\newcolumntype{P}[1]{>{\centering\arraybackslash}p{#1}}
\newcolumntype{M}[1]{>{\centering\arraybackslash}m{#1}}
\usepackage{stackengine}

\usepackage[utf8]{inputenc}
\usepackage[font=small]{caption} % 修复 caption 包的 small 选项问题
\usepackage{amsthm}
\usepackage{booktabs}
\usepackage[switch]{lineno}
\usepackage{adjustbox}

\ExplSyntaxOn
\cs_gset:Npn \__first_footerline:
{
  \group_begin:
  \small
  \sffamily
  \ifnum\theblind>0\relax
  \else
    \__short_authors: :~
  \fi
  {\rmfamily\itshape Under~review}
  \group_end:
}
\ExplSyntaxOff

\begin{document}
\let\WriteBookmarks\relax
\def\floatpagepagefraction{1}
\def\textpagefraction{.001}

\title[mode = title]{PRISM-Net: Patient-specific reference-guided inter-breast symmetry matching for three-class breast DCE-MRI classification}

\author[1,2]{Boya Zhang}[
  orcid=0009-0004-2081-0964
]
\fnmark[1]
\ead{boyazhang@mail.nankai.edu.cn}
\credit{Conceptualization, Methodology, Investigation, Data curation, Validation, Formal analysis, Funding acquisition, Writing -- original draft}

\author[4]{Shuaiwen Zhou}[
  orcid=0009-0008-1245-5113
]
\fnmark[1]
\ead{qdzsw666@bupt.edu.cn}
\credit{Conceptualization, Methodology, Software, Investigation, Formal analysis, Validation, Visualization, Data curation, Writing -- original draft}

\author[6,2]{Di Kong}
\ead{kd24@mails.tsinghua.edu.cn}
\credit{Methodology, Supervision, Conceptualization, Review and editing}

\author[5]{Mingxu Wang}
\ead{wmxwork999@163.com}
\credit{Data curation, Review and editing}

\author[7,2]{Wenbiao Du}
\ead{duwenbiao@bit.edu.cn}
\credit{Data curation, Review and editing}

\author[8,2]{Yiman Zhong}
\ead{yeeman@buaa.edu.cn}
\credit{Data curation, Review and editing}

\author[1,2]{Yuexin Duan}
\ead{duanyuexiner@163.com}
\credit{Data curation, Review and editing}

\author[1,2]{Xiawei Yue}
\ead{yxw@mail.nankai.edu.cn}
\credit{Data curation, Review and editing}

\author[5]{Liuquan Cheng}
\cormark[1]
\ead{13910209982@139.com}
\credit{Supervision, Resources, Project administration, Conceptualization, Review and editing}

\author[1,3]{Xiru Li}
\cormark[2]
\ead{2468li@sina.com}
\credit{Supervision, Funding acquisition, Resources, Project administration, Review and editing}

%% Equal contribution
\fntext[fn1]{These authors contributed equally to this work.}

%% Corresponding authors
\cortext[cor1]{Corresponding author. E-mail: 13910209982@139.com}
\cortext[cor2]{Corresponding author. E-mail: 2468li@sina.com}

%% --- Affiliations ---

\affiliation[1]{
organization={Nankai University},
addressline={No. 94 Weijin Road, Nankai District},
city={Tianjin},
postcode={300071},
country={China}
}

\affiliation[2]{
organization={Zhongguancun Academy},
city={Beijing},
country={China}
}

\affiliation[3]{
organization={The First Medical Center of Chinese PLA General Hospital},
city={Beijing},
country={China}
}
\affiliation[4]{
organization={Beijing University of Posts and Telecommunications},
city={Beijing},
country={China}
}

\affiliation[5]{
organization={The Six Medical Center of Chinese PLA General Hospital},
city={Beijing},
country={China}
}

\affiliation[6]{
organization={Tsinghua University},
city={Beijing},
country={China}
}

\affiliation[7]{
organization={Beijing Institute of Technology},
city={Beijing},
country={China}
}

\affiliation[8]{
organization={Beihang University},
city={Beijing},
country={China}
}

\begin{abstract}
Breast DCE-MRI classification of breast sides as no lesion, benign, or malignant is complicated by substantial patient-specific variation in breast anatomy, tissue composition, and physiological enhancement. Although radiologists routinely use the contralateral breast as an internal reference, computational bilateral comparison remains challenging because corresponding contralateral regions are not anatomically identical and bilateral differences are not necessarily pathological. We propose PRISM-Net, a registration-free bilateral framework that treats the contralateral breast as an approximate patient-specific reference. PRISM-Net learns complementary target-appearance and reference-conditioned deviation representations. Adaptive patch-level matching constructs content-dependent contralateral references in a shared feature space without voxel-wise registration, while local-contrast reweighting emphasizes focal deviations over diffuse bilateral mismatch. The resulting deviation features are integrated with intrinsic target-side appearance through residual cross-attention and aggregated across slices for breast-side classification. On ODELIA, PRISM-Net achieved Macro AUC, Micro AUC, and quadratic weighted kappa values of $84.11 \pm 2.33$, $90.64 \pm 1.61$, and $60.94 \pm 5.64$ on the in-distribution test set, and $68.51 \pm 4.54$, $80.74 \pm 2.68$, and $43.45 \pm 7.10$ on the held-out institution, respectively, yielding the highest point estimates among the evaluated methods across the primary metrics. PRISM-Net also demonstrated favorable performance on an independent institutional cohort and background-complexity subsets, while ablation experiments supported the contributions of adaptive reference matching and focal deviation modeling. These findings support patient-specific contralateral reference learning as a clinically grounded strategy for breast DCE-MRI classification under heterogeneous background conditions.
\end{abstract}

% \noindent{\Large Highlights}
% \vspace{0.5\baselineskip}

% \noindent\textbf{PRISM-Net: Patient-specific reference-guided inter-breast symmetry matching for three-class breast DCE-MRI classification}
% \vspace{0.5\baselineskip}

% \noindent Boya Zhang, Shuaiwen Zhou, Di Kong, Mingxu Wang, Wenbiao Du, Yiman Zhong, Yuexin Duan, Xiawei Yue, Liuquan Cheng and Xiru Li

% \begin{itemize}
% \item PRISM-Net proposes a radiology-guided bilateral framework for breast DCE-MRI classification.

% \item Adaptive patch matching establishes patient-specific contralateral references without anatomical registration.

% \item Models bilateral deviations to distinguish informative asymmetry from background variation.

% \item Improves no-lesion, benign,malignant three class discrimination and ordinal agreement across cohorts.

% \item Demonstrates robustness under complex BPE and FGT patterns.
% \end{itemize}


%% --- Short metadata (for running headers) ---
\shorttitle{PRISM-Net for Breast DCE-MRI}
\shortauthors{Zhang et~al.}
\maketitle

% Main text
\section{Introduction}
\label{introduction}
Breast cancer remains a leading cause of cancer-related mortality among women worldwide \citep{bray2024, siegel}. Dynamic contrast-enhanced breast magnetic resonance imaging (DCE-MRI) is highly sensitive for breast cancer detection and is widely used for screening, diagnosis, staging, and treatment assessment \citep{abdullah2025,kuhl2024,hirsch2025}. Nevertheless, its interpretation remains challenging because breast anatomy, tissue composition, physiological activity, and imaging appearance vary substantially across patients. These patient-specific background signals may resemble suspicious findings or reduce the conspicuity of true abnormalities, contributing to both over-calling and under-calling \citep{giess2014,ray2018,bechynaBaltzer2025}. Separating lesion-related evidence from background-signal interference is therefore a fundamental problem in breast DCE-MRI interpretation.

\begin{figure*}[t]
\centering
\includegraphics[
  width=0.98\textwidth,
  trim={0.8cm 0.6cm 0.8cm 0.6cm},
  clip,
  keepaspectratio
]{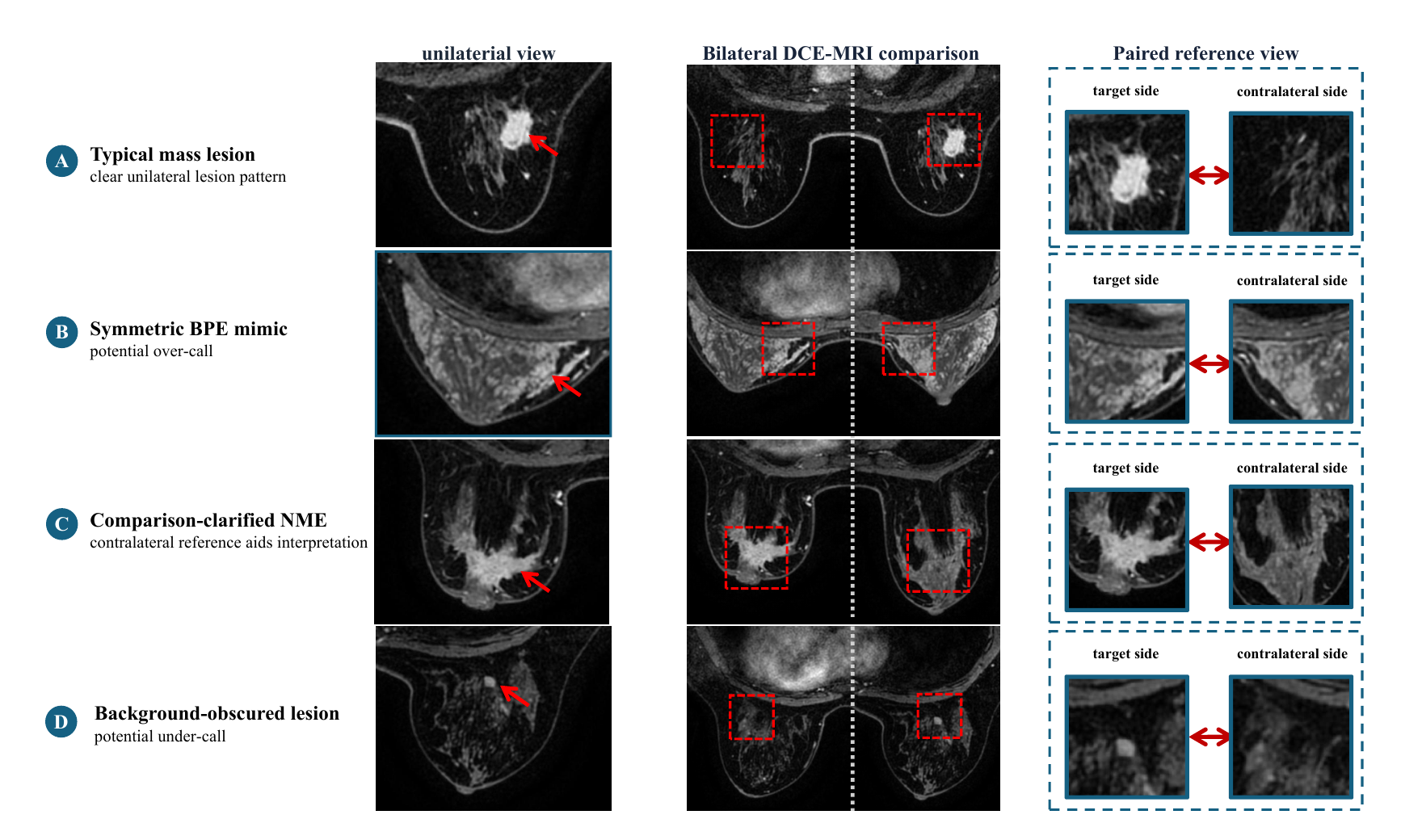}
\caption{Representative clinical examples demonstrating bilateral comparison in breast DCE-MRI. Each row presents unilateral and bilateral views, with paired target-side and contralateral reference crops.
(A) Typical mass lesion in a $56$-year-old woman. An irregular, mildly lobulated right upper-outer breast mass showed asymmetric heterogeneous enhancement. The lesion was assessed as BI-RADS $5$, and biopsy confirmed invasive carcinoma.
(B) Symmetric BPE mimic in a $23$-year-old woman with dense fibroglandular tissue and moderate symmetric BPE. Suspicious enhancement on the unilateral view had a corresponding contralateral pattern, supporting symmetric BPE; both breasts were assessed as BI-RADS $1$(no suspicious enhancement).
(C) Comparison-clarified NME in a $44$-year-old woman. Asymmetric progressive enhancement in the left upper breast comprised NME with multiple small satellite masses. Bilateral comparison showed no corresponding contralateral enhancement, clarifying the finding as NME. The lesion was assessed as BI-RADS $5$, and pathology confirmed invasive carcinoma.
(D) Background-obscured lesion in a $39$-year-old woman with dense fibroglandular tissue and moderate symmetric BPE. A $10$-mm right upper-inner breast mass could be overlooked among surrounding enhancement. The absent contralateral counterpart facilitated detection; the right breast was assessed as BI-RADS $3$, and pathology confirmed fibroadenoma.
BPE, background parenchymal enhancement; DCE-MRI, dynamic contrast-enhanced magnetic resonance imaging; NME, non-mass enhancement.
}
\label{fig:clinical}
\end{figure*}

Bilateral comparison provides a clinically intuitive strategy for reducing background-signal interference in breast DCE-MRI. Also, previous computer-aided and multiparametric imaging studies indicated that bilateral asymmetry and contralateral healthy-tissue biomarkers can provide complementary diagnostic information \citep{yang2014,leithner2019}. This prior is particularly informative in DCE-MRI, where contrast uptake occurs in both lesions and normal fibroglandular tissue. In the BI-RADS lexicon, fibroglandular tissue (FGT) characterizes the amount and distribution of glandular tissue, whereas background parenchymal enhancement (BPE) describes its physiological enhancement after contrast administration. FGT and BPE therefore provide complementary descriptors of background-signal complexity, which has been associated with reduced diagnostic performance \citep{bechynaBaltzer2025,acrBiradsAtlas2013,liao2020,bahl2025}. As illustrated in Fig.~\ref{fig:clinical}, using the contralateral breast as a patient-specific internal reference can help dismiss bilaterally similar mimics, clarify abnormalities without a contralateral counterpart, and reveal lesions obscured by surrounding background. The diagnostic value of bilateral comparison therefore lies not in asymmetry detection alone, but in leveraging shared physiological characteristics to approximate a patient-specific baseline for interpreting suspicious deviations.

Although deep learning studies in mammography and digital breast tomosynthesis have demonstrated the diagnostic value of bilateral information \citep{shimokawa2023bilad,wang2023disasymnet,zeng2025,pelluet2025}, the clinical prior of bilateral comparison remains underused in deep learning models for breast MRI \citep{hao2020,logullo2024,gravina2021,gravina2024,oviedo2025cancer,zhang2023maskrcnnresnet}. At the modeling level, many methods remain lesion-centric, single-sided, or only implicitly bilateral. Fine-grained bilateral modeling requires correspondence between target regions and contralateral tissue. Conventional correspondence estimation often relies on deformable registration, which remains challenging in breast DCE-MRI because of nonrigid breast motion and enhancement-related temporal intensity changes \citep{zheng2015,truhn2019,antropova2019,hizukuri2021,sun2023}. At the task level, most breast MRI classification models have been developed in lesion-enriched binary settings. However, as breast MRI expands across screening and diagnostic applications, examinations without suspicious findings constitute a clinically relevant part of the case spectrum \citep{marcon2024,verburg2022}. The recently released multicenter ODELIA resource provides breast-side labels for \emph{no-lesion}, \emph{benign}, and \emph{malignant} findings, yet studies in this setting remain scarce \citep{mullerfranzes2025odelia,odeliaChallenge2025}. Taken together, the central gap is the lack of an end-to-end breast MRI framework that jointly learns registration-free, spatially adaptive cross-breast correspondence, treats the contralateral breast as an approximate patient specific reference rather than a normal template, and distinguishes diagnostically relevant focal deviations from diffuse or bilaterally shared variation while preserving intrinsic target side appearance. This gap is particularly consequential for breast side three class classification, where the model must distinguish \emph{no lesion}, \emph{benign}, and \emph{malignant} findings under heterogeneous anatomical and physiological backgrounds without assuming that a target lesion has already been identified. 

To address this gap, we formulate bilateral DCE-MRI analysis as patient-specific reference-guided interpretation, in which the contralateral breast provides approximate internal context for identifying localized deviations from shared anatomical and physiological background patterns. We therefore propose PRISM-Net, a patient-specific reference-guided inter-breast symmetry matching for three-class breast DCE-MRI classification without explicit anatomical registration. PRISM-Net encodes the target and contralateral breasts using a shared-weight backbone to establish a common feature space. Bidirectional patch-level cross-breast matching then retrieves an adaptive contralateral reference for each target region under anatomical variability. The model integrates target features, matched references, directional differences, and cross-breast interactions to construct reference-conditioned representations of bilateral deviations. A local-contrast reweighting mechanism further suppresses diffuse or bilaterally shared variation while emphasizing diagnostically relevant focal asymmetries. The resulting deviation evidence is fused with intrinsic target-side appearance and aggregated across slices to classify each breast as no-lesion, benign, or malignant. By converting clinical bilateral comparison into a learnable diagnostic prior, PRISM-Net provides a unified framework for patient-specific reference modeling and breast-side classification..

We evaluate PRISM-Net across the public multicenter ODELIA dataset \citep{odeliaChallenge2025} and independent institutional DCE-MRI cohorts, all with breast-side labels for \emph{no-lesion}, \emph{benign}, and \emph{malignant} findings. Bilateral no-lesion examinations supported by clinical BI-RADS assessments are retained to represent lesion-free settings, while dedicated BPE- and FGT-complexity cohorts capture challenging backgrounds. Together, this multi-cohort design provides a clinically oriented assessment of three-class performance, cross-cohort generalizability, and robustness to background-signal complexity.

The contributions of this work are as follows.
\begin{enumerate}
  \item We formulate breast-side DCE-MRI classification as a patient-specific bilateral reference learning problem. The contralateral breast is treated as an approximate internal reference for distinguishing shared anatomical and physiological background patterns from localized lesion-related deviations across \emph{no-lesion}, \emph{benign}, and \emph{malignant} findings.

  \item We propose PRISM-Net, a registration-free bilateral framework that learns complementary target-appearance and reference-conditioned deviation representations. Its adaptive patch-level matching mechanism constructs content-dependent contralateral references in a shared feature space, avoiding fixed mirrored correspondence and deformable registration.

  \item We introduce local-contrast focal deviation modeling and appearance-preserving fusion. The former emphasizes spatially localized target-to-reference deviations over diffuse bilateral mismatch, whereas the latter integrates these cues with intrinsic target-side appearance through a residual pathway for volumetric classification.

  \item We evaluate PRISM-Net on the public multicenter ODELIA dataset, a held-out external ODELIA institution, and independent institutional cohorts including BPE- and FGT-complexity subsets. The results show favorable discrimination and ordinal agreement relative to the evaluated baselines, and ablation analyses support the contributions of adaptive contralateral matching and focal deviation modeling.
\end{enumerate}

%% =============================================================================
\section{Related work}
\label{sec:relatedwork}
%% =============================================================================

\subsection{Deep learning in breast MRI diagnosis}  

Existing deep learning research in breast MRI has predominantly focused on lesion-level or binary diagnostic tasks, commonly using localized lesion regions or unilateral breast inputs.

Deep learning in breast MRI has predominantly focused on lesion-level characterization. CNNs, recurrent networks, projection-based models, and multi-task architectures have been used to distinguish benign from malignant lesions from manually segmented, radiologist-selected, or automatically detected regions. Representative approaches include multiparametric CNN classification, four-dimensional CNN–LSTM aggregation, deep-feature maximum-intensity projections, and joint lesion segmentation and classification \citep{truhn2019,antropova2019,hizukuri2021,hu2021,sun2023}. DAE-CNN and its PBPK-informed extension further incorporate contrast-agent disentanglement and physiologically constrained data augmentation to improve lesion representation under limited training data \citep{gravina2021,gravina2024}. Despite differences in architecture and input representation, these methods generally assume that a target lesion has already been identified. 

Existing breast MRI diagnostic models also predominantly adopt binary task formulations. Lesion-level studies commonly distinguish benign from malignant findings, whereas breast-level or volume-level models usually predict malignant versus non-malignant outcomes \citep{mullerFranzes2025MST,hirsch2025}. Such formulations either exclude breasts without detectable lesions or combine no-lesion and benign cases within a single negative category. The ODELIA dataset and benchmark provide a clinically relevant task setting by assigning no-lesion, benign, and malignant labels to each breast in multicentre DCE-MRI examinations \citep{mullerfranzes2025odelia,odeliaChallenge2025}. Its Medical Slice Transformer baseline aggregates slice-level features from pre-contrast T1-weighted, first post-contrast subtraction, and T2-weighted volumes, establishing breast-side three-class diagnosis as a direct precedent \citep{mullerfranzes2025odelia}.Because the two breasts are treated as separate samples, the model does not explicitly learn cross-breast interactions. 

Taken together, current breast MRI deep-learning methods remain largely shaped by lesion-level inputs and binary label spaces. Model development would benefit from prediction units and label definitions that better reflect clinical diagnostic tasks. Breast-side three-class classification addresses this requirement by jointly distinguishing no-lesion, benign, and malignant states without presuming that a target lesion has already been identified.

\subsection{Bilateral comparison and symmetry-aware learning in breast imaging}

Bilateral comparison is a routine interpretive strategy in breast imaging, allowing local morphology and enhancement to be evaluated against a patient-specific internal reference.  Because the contralateral breast may itself contain abnormalities and bilateral anatomy is affected by natural asymmetry, nonrigid deformation, positioning variation, and heterogeneous background enhancement, bilateral symmetry is an approximate prior and direct voxel-wise correspondence is unreliable \citep{alterson2003,hennessey2014bilateral,jansen2011normal,liao2020,zeng2025}. Early breast MRI studies quantified bilateral symmetry using registration-free handcrafted three-dimensional descriptors or predefined whole-breast kinetic differences, but remained limited to global representations \citep{alterson2003,yang2014}.

In subsequent breast MRI deep learning studies, bilateral information has not been entirely absent, but it was often incorporated implicitly. Zhang et al. proposed a two-stage framework using Mask R-CNN for suspicious lesion detection followed by ResNet50 for malignancy classification in breast MRI \citep{zhang2023maskrcnnresnet}. In the detection stage, the model used pre-contrast images and subtraction images of the left and right breasts as inputs, allowing breast symmetry to be considered within the whole-breast context. However, the method did not explicitly establish correspondence between the two breasts, nor did it construct asymmetry features, a contralateral reference branch, bilateral attention, or a dedicated bilateral fusion module. The subsequent classification stage mainly focused on local candidate regions detected by Mask R-CNN and used DCE parametric maps for lesion characterization. Therefore, this type of work suggested that breast MRI deep learning had recognized the importance of bilateral information, but such information was mainly incorporated as global contextual input. Within BreastMRI-FCDD, which performed breast-level binary cancer detection on DCE-MRI subtraction maximum intensity projections, Oviedo et al. \citep{oviedo2025cancer} further investigated FCDD-Symmetric, in which the contralateral breast representation defined the normal-class center for anomaly scoring. However, its projection-level, center-based formulation did not explicitly establish spatially adaptive correspondence between bilateral three-dimensional feature volumes or use spatially matched contralateral features for background signal baseline normalization.  

In mammography, digital breast tomosynthesis, and other breast imaging modalities, bilateral information has been more explicitly embedded into deep learning models. Earlier studies introduced distortion-insensitive, logic-guided bilateral comparison and paired asymmetry modeling for cancer localization and classification, while BilAD subsequently extended bilateral asymmetry learning to DBT \citep{shimokawa2023bilad,liu2019contrasted,guan2020asymmetric}. DisAsymNet attempted to disentangle asymmetric abnormalities from symmetric breast structures in bilateral mammograms, reconstructing abnormality-removed symmetric images to improve interpretability \citep{wang2023disasymnet}. A recent bilateral information-guided Vision Transformer framework further proposed a registration-free strategy for mammography, directly concatenating bilateral images and introducing soft spatial prompts to model cross-breast structural differences without explicit image registration \citep{zeng2025}. BiGAM-Net also used asymmetry-aware fusion and inspectable gating mechanisms to quantify model reliance on intrinsic lesion features and contralateral asymmetry evidence in paired breast imaging \citep{haque2026bigamnet}. These studies collectively suggest that bilateral information can serve as a structural medical prior to improve diagnostic performance and interpretability.

These studies support the value of explicit bilateral reasoning, while their two-dimensional or quasi-two-dimensional formulations provide limited coverage of the deformable volumetric anatomy and heterogeneous enhancement of breast DCE-MRI. Registration-free, spatially adaptive cross-breast matching coupled with patient-specific background normalization therefore remains the central methodological gap.

%% =============================================================================
\section{Methods}
\label{sec:methods}
\begin{figure*}[t]
    \centering
    \includegraphics[
        width=0.95\textwidth,
    ]{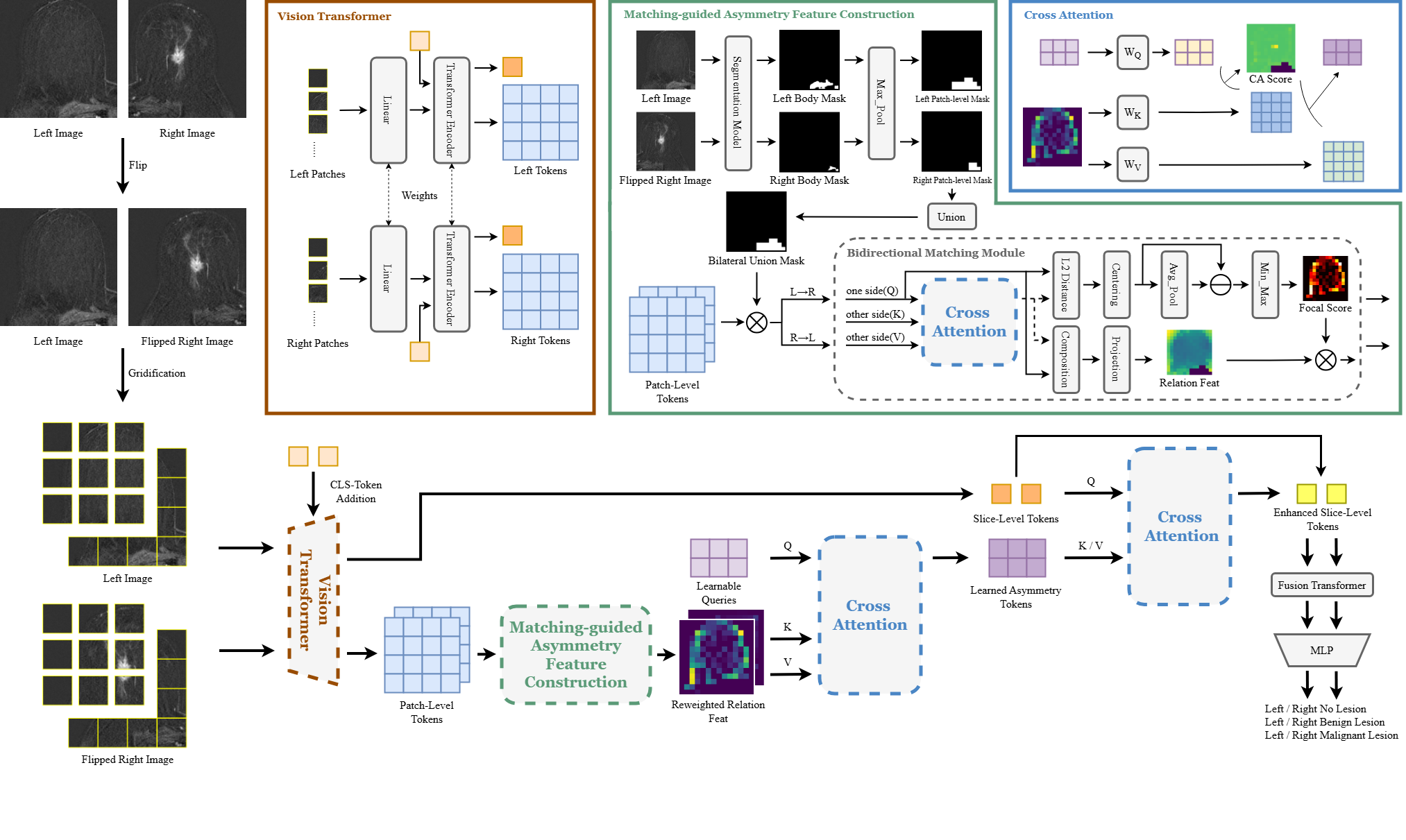}
    \caption{Overall architecture of the proposed PRISM-Net for breast DCE-MRI classification.}
    \label{fig:architecture}
\end{figure*}

%% =============================================================================
PRISM-Net translates radiological bilateral comparison into a registration-free framework that conditions each target breast on an adaptively matched contralateral reference, as illustrated in Fig.~\ref{fig:architecture}. We first formulate bilateral breast-side classification and introduce the clinical rationale for patient-specific reference learning in Sec.~\ref{sec:bilateral_formulation}. We then establish a shared bilateral representation space in Sec.~\ref{sec:shared_representation} and construct adaptive patch level contralateral references without voxel-wise registration in Sec.~\ref{sec:reference_matching}. Next, Sec.~\ref{sec:focal_modeling} introduces focal bilateral-deviation modeling to attenuate diffuse mismatch and summarize localized reference-conditioned evidence. Finally, Sec.~\ref{sec:volumetric_fusion} integrates this evidence with target-side appearance and aggregates it across slices for breast-side three-class classification.

\subsection{Bilateral problem formulation}
\label{sec:bilateral_formulation}

PRISM-Net is motivated by the clinical use of bilateral comparison in breast DCE-MRI. Rather than interpreting a target-side finding solely from its absolute appearance, radiologists compare it with contralateral tissue to determine whether it represents a meaningful deviation from patient-specific anatomical and physiological background patterns. Because the contralateral breast is neither anatomically identical to the target breast nor guaranteed to be lesion-free, we treat it as an approximate patient-specific reference rather than a normal template.

Operationalizing this reference-based interpretation raises three modeling challenges. First, corresponding bilateral regions are not necessarily located at identical mirrored coordinates because of nonrigid breast deformation, patient positioning, and anatomical asymmetry, making direct left-right subtraction unreliable. Second, bilateral mismatch may arise from moderate/marked BPE, heterogeneous/dense FGT, or other natural asymmetry rather than local pathology. Third, bilateral asymmetry alone is insufficient for three-class diagnosis, because distinguishing no-lesion, benign, and malignant findings also requires intrinsic target-side appearance and cross-slice volumetric context.

Accordingly, PRISM-Net learns two complementary representations for each breast: a target-appearance representation that preserves intrinsic diagnostic features, and a reference-conditioned deviation representation that characterizes how local target features differ from adaptively matched contralateral context.

Formally, let $\mathbf{X}^{L}$ and $\mathbf{X}^{R}$ denote the paired left and right breast DCE-MRI volumes from one examination, with $\mathbf{X}^{L},\mathbf{X}^{R}\in\mathbb{R}^{D\times H\times W}$, where $D$ is the number of axial slices and $H$ and $W$ are the in-plane spatial dimensions. Let $\mathcal{S}=\{L,R\}$ denote the set of breast sides. For each side $a\in\mathcal{S}$, the diagnostic label is $y^{a}\in\{0,1,2\}$, corresponding to no-lesion, benign, and malignant findings, respectively. Given the paired bilateral volumes, PRISM-Net learns a function $f_{\theta}$ that jointly predicts the side-specific class probability distributions $(\boldsymbol{\pi}^{L},\boldsymbol{\pi}^{R})=f_{\theta} (\mathbf{X}^{L},\mathbf{X}^{R})$, where $\boldsymbol{\pi}^{a}\in[0,1]^3$ and $\sum_{c=0}^{2}\pi_{c}^{a}=1$ for each $a\in\mathcal{S}$.

\subsection{Shared bilateral representation}
\label{sec:shared_representation}

To make the approximate contralateral reference computationally usable, the bilateral volumes must first be mapped into a comparable semantic space. PRISM-Net horizontally flips the right breast volume, denoted by $\bar{\mathbf{X}}^{R}=\operatorname{Flip}(\mathbf{X}^{R})$, to place both breasts in a common orientation. This operation provides only a coarse spatial prior and does not assume voxel-wise correspondence. The left volume $\mathbf{X}^{L}$ and the flipped right volume $\bar{\mathbf{X}}^{R}$ are then encoded slice-wise using a shared DINOv3 ViT-S/16 backbone~\citep{simeoni2025dinov3}. Weight sharing ensures that bilateral features are represented in the same semantic space, thereby enabling subsequent feature-space matching.

For the $d$-th slice and breast side $a\in\mathcal{S}$, the shared encoder produces a global slice token and a matrix of local patch tokens:
\begin{equation}
\mathbf{s}_{d}^{a}\in\mathbb{R}^{C},
\qquad
\mathbf{P}_{d}^{a}\in\mathbb{R}^{N\times C},
\end{equation}
where $\mathbf{s}_{d}^{a}$ preserves the global target-side appearance, $\mathbf{P}_{d}^{a}$ contains the corresponding local patch representations, $C$ is the feature dimension, and $N$ is the number of image patches. The patch-level representations are subsequently used to construct adaptive contralateral references, whereas the global slice tokens are retained for appearance-preserving classification.

\subsection{Registration-free adaptive contralateral reference matching}
\label{sec:reference_matching}

With bilateral features represented in a common semantic space, the next step is to construct a patient-specific reference for each target patch. Because identical mirrored coordinates do not guarantee anatomical correspondence, direct patch-wise subtraction may confound pathological differences with variations in breast shape, positioning, and deformation. PRISM-Net therefore allows each target patch to retrieve an adaptive contralateral reference through soft feature-space matching within the paired axial slice, without requiring voxel-wise registration.

Before matching, unreliable regions are excluded using binary masks. Let $\mathbf{M}^{L},\mathbf{M}^{R}\in\{0,1\}^{D\times H\times W}$ denote the exclusion masks corresponding to the original bilateral volumes, where $1$ indicates a region excluded from matching. The right-side mask is flipped together with the right breast volume, $\bar{\mathbf{M}}^{R}=\operatorname{Flip}(\mathbf{M}^{R})$, and the bilateral exclusion mask in the common orientation is defined as $\mathbf{M}^{U}=\mathbf{M}^{L}\vee\bar{\mathbf{M}}^{R}$. For slice $d$, the valid patch set is

\begin{equation}
\mathcal{V}_{d}
=
\left\{
i\;\middle|\;
\max_{(u,v)\in\mathcal{O}_{i}}
M^{U}_{d,u,v}=0
\right\},
\end{equation}

where $\mathcal{O}_{i}$ denotes the spatial support of patch $i$. This filtering ensures that bilateral matching is performed only between patches that are reliable on both sides.

For one matching direction, let $a\in\mathcal{S}$ denote the target side and let $b\in\mathcal{S}$, $b\neq a$, denote the contralateral reference side. Based on the valid patch-token matrices, the matching weights and adaptive contralateral references are computed as

\begin{equation}
\begin{aligned}
\mathbf{A}_{d}^{a\leftarrow b}
&=
\operatorname{softmax}
\left(
\frac{
(\mathbf{P}_{d,\mathcal{V}_{d}}^{a}\mathbf{W}_{Q})
(\mathbf{P}_{d,\mathcal{V}_{d}}^{b}\mathbf{W}_{K})^{\top}
}{
\sqrt{d_{k}}
}
\right),\\
\mathbf{R}_{d}^{a}
&=
\mathbf{A}_{d}^{a\leftarrow b}
(\mathbf{P}_{d,\mathcal{V}_{d}}^{b}\mathbf{W}_{V}),
\end{aligned}
\label{eq:adaptive_matching}
\end{equation}

where $d_{k}$ is the projected query--key dimension and $\mathbf{W}_{Q}$, $\mathbf{W}_{K}$, and $\mathbf{W}_{V}$ are learnable projection matrices. The softmax operation is applied over the valid contralateral patches, so each row of $\mathbf{A}_{d}^{a\leftarrow b}$ describes the correspondence distribution between one target patch and the candidate reference patches. Consequently, each row $\mathbf{r}_{d,i}^{a}$ of $\mathbf{R}_{d}^{a}$ represents a soft, content-adaptive contralateral reference for the corresponding target token $\mathbf{p}_{d,i}^{a}$, rather than the feature at a fixed mirrored location. The same matching operation is applied in both directions with shared parameters. These adaptively matched reference pairs provide the basis for bilateral relation modeling. However, their discrepancies are not uniformly diagnostic.

\subsection{Focal bilateral-deviation modeling}
\label{sec:focal_modeling}

A large target-to-reference difference does not necessarily signify pathology, as it may arise from natural asymmetry in breast morphology, FGT distribution, asymmetric BPE, hormonal or lactational influences, or inflammatory and edematous changes. Therefore, after obtaining the adaptive contralateral reference, PRISM-Net preserves both the target and reference representations and explicitly encodes their bilateral relationship.

For each valid patch $i\in\mathcal{V}_{d}$, the bilateral relation representation is constructed as

\begin{equation}
\begin{aligned}
\mathbf{c}_{d,i}^{a}
&=
\operatorname{Concat}
\left(
\mathbf{p}_{d,i}^{a},
\mathbf{r}_{d,i}^{a},
\mathbf{p}_{d,i}^{a}-\mathbf{r}_{d,i}^{a},
\mathbf{p}_{d,i}^{a}\odot\mathbf{r}_{d,i}^{a}
\right),\\
\mathbf{g}_{d,i}^{a}
&=
\phi\left(\mathbf{c}_{d,i}^{a}\right),
\end{aligned}
\label{eq:bilateral_relation}
\end{equation}

where $\mathbf{p}_{d,i}^{a}$ preserves target-side appearance, $\mathbf{r}_{d,i}^{a}$ provides patient-specific contralateral context, their difference encodes the directional target-to-reference deviation, and their element-wise product captures feature-wise bilateral interaction. The projection layer $\phi(\cdot)$ maps the concatenated representation to a $C$-dimensional relation feature $\mathbf{g}_{d,i}^{a}$.

Although the relation representation captures bilateral differences, diffuse mismatch may reflect background variation rather than focal pathology. PRISM-Net therefore measures how strongly each patch-level discrepancy stands out from its local spatial neighborhood. Let $\mathcal{N}_{d}(i)=\mathcal{N}(i)\cap\mathcal{V}_{d}$ denote the valid patches within the $3\times3$ neighborhood of patch $i$. The local-contrast asymmetry score and reweighted relation feature are computed as

\begin{equation}
\begin{aligned}
A_{d,i}^{a}
&=
\left\|
\mathbf{p}_{d,i}^{a}-\mathbf{r}_{d,i}^{a}
\right\|_{2},\\
\bar{A}_{d,i}^{a}
&=
\frac{1}{|\mathcal{N}_{d}(i)|}
\sum_{j\in\mathcal{N}_{d}(i)} A_{d,j}^{a},\\
F_{d,i}^{a}
&=
\operatorname{ReLU}
\left(A_{d,i}^{a}-\bar{A}_{d,i}^{a}\right),\\
\widetilde{F}_{d,i}^{a}
&=
\frac{
F_{d,i}^{a}-\min_{j\in\mathcal{V}_{d}}F_{d,j}^{a}
}{
\max_{j\in\mathcal{V}_{d}}F_{d,j}^{a}
-\min_{j\in\mathcal{V}_{d}}F_{d,j}^{a}
+\epsilon
},\\
\widehat{\mathbf{g}}_{d,i}^{a}
&=
\widetilde{F}_{d,i}^{a}\mathbf{g}_{d,i}^{a}.
\end{aligned}
\label{eq:focal_reweighting}
\end{equation}

This local-contrast formulation emphasizes discrepancies that are salient relative to their surroundings while attenuating spatially diffuse bilateral mismatch.

Because diagnostically relevant deviations may be sparse or distributed across multiple regions, direct global averaging could dilute their contribution. We therefore stack the reweighted relation features as $\widehat{\mathbf{G}}_{d}^{a}\in
\mathbb{R}^{|\mathcal{V}_{d}|\times C}$ and summarize them using a shared set of learnable asymmetry queries $\mathbf{Q}_{A}\in\mathbb{R}^{K\times C}$:

\begin{equation}
\mathbf{T}_{d}^{a}
=
\operatorname{CA}
\left(
\mathbf{Q}_{A},
\widehat{\mathbf{G}}_{d}^{a},
\widehat{\mathbf{G}}_{d}^{a}
\right),
\qquad
\mathbf{T}_{d}^{a}\in\mathbb{R}^{K\times C}.
\label{eq:asymmetry_tokens}
\end{equation}

The complete relation-modeling module is applied bidirectionally with shared parameters. The resulting tokens remain side-specific because the target and contralateral reference roles are reversed in the two matching directions. The resulting side-specific asymmetry tokens summarize focal reference-conditioned deviations, but they do not replace the intrinsic appearance of the target breast. We therefore integrate them with the global target-side slice representations for final volumetric classification.

\subsection{Appearance-preserving volumetric fusion and classification}
\label{sec:volumetric_fusion}

The preceding module produces side-specific tokens that summarize focal bilateral deviations. For diagnosis, however, this information should complement rather than replace intrinsic target-side appearance, because symmetric findings are not necessarily normal and asymmetric findings are not necessarily pathological. Moreover, the contralateral breast may itself contain abnormalities. PRISM-Net therefore retains the target-side slice token as the primary appearance representation and introduces the bilateral-deviation tokens through a residual cross-attention pathway.

For side $a\in\mathcal{S}$ and slice $d$, the reference-conditioned slice representation is computed as

\begin{equation}
\widetilde{\mathbf{s}}_{d}^{a}
=
\mathbf{s}_{d}^{a}
+
\lambda
\operatorname{CA}
\left(
\mathbf{s}_{d}^{a},
\mathbf{T}_{d}^{a},
\mathbf{T}_{d}^{a}
\right),
\label{eq:residual_fusion}
\end{equation}

where $\lambda$ is a fixed residual scaling hyperparameter that controls the contribution of the bilateral-deviation information. The residual scaling factor $\lambda$ was fixed at $0.8$ throughout training and evaluation. This residual formulation preserves the intrinsic target-side appearance while allowing the slice representation to be selectively conditioned on patient-specific contralateral context.

Because diagnostically relevant evidence may extend across multiple slices, the reference-conditioned slice representations are subsequently aggregated to capture volumetric context. To preserve slice order, a positional embedding $\mathbf{e}_{d}\in\mathbb{R}^{C}$ is added to each reference-conditioned slice token. A learnable fusion token $\mathbf{u}\in\mathbb{R}^{C}$ is then appended to the resulting sequence, and a slice-level Transformer produces the side-specific volumetric representation:

\begin{equation}
\mathbf{z}^{a}
=
\operatorname{Transformer}
\left(
[
\widetilde{\mathbf{s}}_{1}^{a}+\mathbf{e}_{1},
\ldots,
\widetilde{\mathbf{s}}_{D}^{a}+\mathbf{e}_{D},
\mathbf{u}
]
\right)_{\mathbf{u}},
\label{eq:slice_fusion}
\end{equation}

where $(\cdot)_{\mathbf{u}}$ denotes the output representation associated with the fusion token. The slice positional embeddings enable the Transformer to distinguish slice locations and model ordered inter-slice context. The same slice-level Transformer is applied to both breast sides with shared parameters.

Finally, a shared MLP classifier $h(\cdot)$ maps the volumetric representation to side-specific logits $\boldsymbol{\ell}^{a}=h(\mathbf{z}^{a})\in\mathbb{R}^{3}$, and the corresponding class probability distribution is $\boldsymbol{\pi}^{a}=\operatorname{softmax}(\boldsymbol{\ell}^{a})$. The classifier parameters are shared between the left and right breasts. PRISM-Net is trained using the summed side-level cross-entropy loss:

\begin{equation}
\mathcal{L}
=
\sum_{a\in\mathcal{S}}
\mathcal{L}_{\mathrm{CE}}
\left(
\boldsymbol{\ell}^{a},
y^{a}
\right).
\label{eq:classification_loss}
\end{equation}

Thus, the left and right breast labels provide separate side-level supervision within a single bilateral forward pass.

%% =============================================================================
\section{Experiments and results}
\label{sec:experiments}
%% =============================================================================

% TODO: Datasets, baselines, metrics, test set analyses.
\subsection{Dataset and cohort characteristics  }
\begin{figure*}[t]
    \centering
    \includegraphics[
        width=0.95\textwidth,
    ]{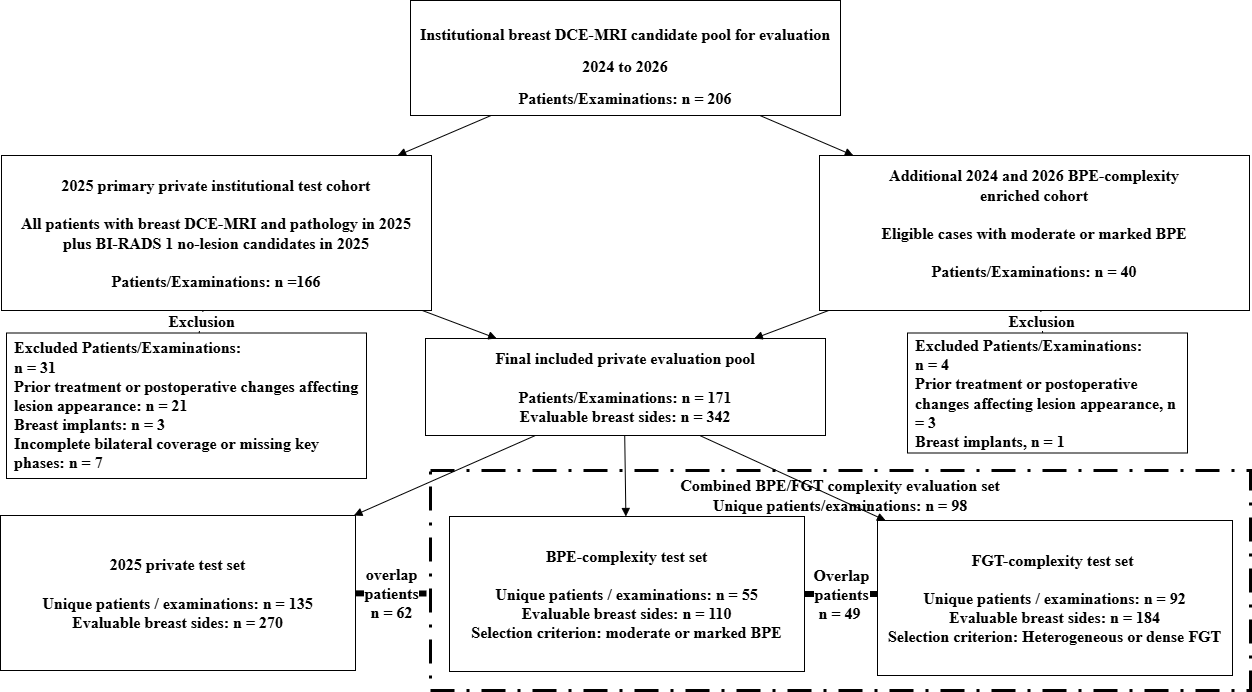}
    \caption{\textbf{Flowchart of institutional patient selection and dataset construction.} The candidate pool comprised $206$ patients, each contributing one bilateral DCE-MRI examination. It included a primary $2025$ institutional cohort of $166$ patients and an additional BPE-enriched cohort of $40$ patients from $2024$ and $2026$, and FGT-complexity membership was subsequently assigned within the final private evaluation pool according to FGT level. After the specified exclusions, the final private evaluation pool comprised $171$ patients and $342$ evaluable breast sides. Three partially overlapping evaluation sets were defined: the $2025$ private test set with $135$ patients and $270$ breast sides, the BPE-complexity set with $55$ patients and $110$ breast sides, and the FGT-complexity set with $92$ patients and $184$ breast sides. The dashed enclosure denotes the union of the two complexity sets, comprising $98$ patients. Among them, $62$ were also included in the $2025$ private test set, and $49$ satisfied both complexity criteria. BPE complexity was defined as moderate or marked BPE, whereas FGT complexity was defined as heterogeneous or dense FGT. DCE-MRI, dynamic contrast-enhanced magnetic resonance imaging; BPE, background parenchymal enhancement; FGT, fibroglandular tissue.}
    \label{fig:patient cohort}
\end{figure*}

%%todo:characteristic
Model development used the public ODELIA breast DCE-MRI dataset. Evaluation was performed using held-out ODELIA data and an independent institutional dataset from the General Hospital of the People's Liberation Army. The selection workflow for the private institutional set is shown in Fig.~\ref{fig:patient cohort}.

\textbf{Public ODELIA dataset.}
The public ODELIA dataset comprised $741$ examinations from $741$ women acquired at six European medical centers between December $2006$ and May $2024$ \citep{odeliaChallenge2025}. Each breast was independently labeled as no-lesion, benign, or malignant. $291$ examinations exhibited no lesions, $146$ examinations were diagnosed with benign but no malignant lesions, and $304$ examinations had malignant lesions. In total, the dataset includes $978$ breasts without lesions, $195$ with benign lesions, and $309$ with a malignant lesion. The original training (n = $408$) and validation (n = $103$) partitions were retained, while the test partition was split into an internal set (n = $130$) and a center-held-out set (n = $100$). 

\textbf{Private institutional dataset.}
An independent institutional cohort was retrospectively assembled to evaluate generalization and performance under complex background conditions. The primary candidate cohort was drawn from the $2025$ institutional archive. As examinations with moderate or marked BPE were underrepresented in this cohort, additional candidates meeting this BPE criterion were retrieved from the $2024$ and $2026$ archives. Examinations were reviewed by two breast radiologists and one clinician using DCE-MRI findings, BI-RADS assessments, radiology reports, and pathological results. Examinations were excluded for substantial treatment-related or postoperative changes, breast implants, or missing key DCE-MRI phases. After patient-level deduplication, $171$ examinations remained ($135$ from $2025$ and $36$ BPE-enriched examinations from $2024$ and $2026$), yielding $342$ evaluable breast sides. Overlapping complexity subsets included $55$ examinations with moderate/marked BPE and $92$ with heterogeneous/dense FGT; $49$ met both criteria, as shown in Fig.~\ref{fig:patient cohort}. $79$ examinations exhibited no lesions, $32$ women were diagnosed with benign but no malignant lesions, and $60$ women had malignant lesions. 

\textbf{Breast-side reference standard }
The left and right breasts were treated as separate prediction targets and labeled as no-lesion, benign, or malignant. Labels were assigned using a pathology-prioritized hierarchy. Malignant or benign labels were assigned according to ipsilateral pathological findings. A no-lesion label was assigned when no ipsilateral enhancing lesion or pathological diagnosis was identified. For breasts with multiple findings, the most severe diagnosis determined the final label, with malignancy taking precedence over benignity and no-lesion. In total, the dataset includes $240$ breasts without lesions, $38$ with benign lesions, and $64$ with a malignant lesion. Cohort characteristics and acquisition protocols are summarized in Tables~\ref{tab:private_cohort_characteristics} and Supplementary Table S1.

\begin{table}[t]
\centering
\scriptsize
\setlength{\tabcolsep}{2pt}
\renewcommand{\arraystretch}{1.08}

\begin{threeparttable}

\caption{Demographic, breast background, and lesion characteristics of the
final institutional breast DCE-MRI evaluation pool and its partially
overlapping evaluation sets.}
\label{tab:private_cohort_characteristics}

\begin{tabularx}{\columnwidth}{
    @{}
    >{\raggedright\arraybackslash}X
    *{4}{>{\centering\arraybackslash}p{0.15\columnwidth}}
    @{}
}
\toprule

\textbf{Characteristic}
&
\textbf{\makecell[c]{Final\\pool}}
&
\textbf{\makecell[c]{2025\\set}}
&
\textbf{\makecell[c]{BPE\\set}}
&
\textbf{\makecell[c]{FGT\\set}}
\\

\midrule

Patients/examinations, $n$
& 171 & 135 & 55 & 92 \\

Evaluable breast sides, $n$
& 342 & 270 & 110 & 184 \\

Age, years\tnote{a}
& $47.5 \pm 13.0$
& $48.8 \pm 13.2$
& $42.0 \pm 11.0$
& $44.2 \pm 11.8$ \\

\addlinespace[2pt]
\multicolumn{5}{l}{\textit{\textbf{BPE level}}\tnote{a}} \\

Minimal/mild, $n$ (\%)
& 116 (67.8)
& 116 (85.9)
& 0 (0.0)
& 43 (46.7) \\

Moderate/marked, $n$ (\%)
& 55 (32.2)
& 19 (14.1)
& 55 (100.0)
& 49 (53.3) \\

\addlinespace[2pt]
\multicolumn{5}{l}{\textit{\textbf{FGT level}}\tnote{a}} \\

Fatty/scattered, $n$ (\%)
& 79 (46.2)
& 76 (56.3)
& 6 (10.9)
& 0 (0.0) \\

Heterogeneous/dense, $n$ (\%)
& 92 (53.8)
& 59 (43.7)
& 49 (89.1)
& 92 (100.0) \\

\addlinespace[2pt]
\multicolumn{5}{l}{\textit{\textbf{Lesion laterality}}\tnote{a}} \\

No-lesion, $n$ (\%)
& 79 (46.2)
& 66 (48.9)
& 23 (41.8)
& 22 (23.9) \\

Unilateral, $n$ (\%)
& 82 (48.0)
& 62 (45.9)
& 27 (49.1)
& 64 (69.6) \\

Bilateral, $n$ (\%)
& 10 (5.8)
& 7 (5.2)
& 5 (9.1)
& 6 (6.5) \\

\addlinespace[2pt]
\multicolumn{5}{l}{\textit{\textbf{Patient-level label}}\tnote{a}} \\

No-lesion, $n$ (\%)
& 79 (46.2)
& 66 (48.9)
& 23 (41.8)
& 22 (23.9) \\

Benign, $n$ (\%)
& 32 (18.7)
& 23 (17.0)
& 13 (23.6)
& 28 (30.4) \\

Malignant, $n$ (\%)
& 60 (35.1)
& 46 (34.1)
& 19 (34.5)
& 42 (45.7) \\

\addlinespace[2pt]
\multicolumn{5}{l}{\textit{\textbf{Breast-side label}}\tnote{b}} \\

No-lesion breast sides, $n$ (\%)
& 240 (70.2)
& 194 (71.9)
& 73 (66.4)
& 108 (58.7) \\

Benign breast sides, $n$ (\%)
& 38 (11.1)
& 28 (10.4)
& 17 (15.5)
& 33 (17.9) \\

Malignant breast sides, $n$ (\%)
& 64 (18.7)
& 48 (17.8)
& 20 (18.2)
& 43 (23.4) \\

\bottomrule
\end{tabularx}

\begin{tablenotes}[flushleft]
\scriptsize

\item[a]
Calculated at the patient/examination level. Age is presented as mean
$\pm$ standard deviation, and categorical variables as $n$ (\%).

\item[b]
Calculated at the individual breast-side level.

\item[]
BPE, background parenchymal enhancement;
FGT, fibroglandular tissue;
DCE-MRI, dynamic contrast-enhanced magnetic resonance imaging.

\end{tablenotes}
\end{threeparttable}
\end{table}

\subsection{Experimental setup}

\subsubsection{Data preprocessing}

A subtraction volume was generated from the pre-contrast and first post-contrast images, and each bilateral volume was split at the midline. The right breast was horizontally flipped to provide coarse bilateral correspondence without voxel-wise registration. Each breast was resized or cropped to a fixed shape, independently intensity-normalized, and masked to suppress non-breast regions. Paired breast volumes and masks were input jointly; no lesion-level annotations or manual registration were required.

\subsubsection{Implementation details}

PRISM-Net was implemented in PyTorch and optimized end-to-end using the summed cross-entropy losses of the left and right breasts. The shared DINOv3 ViT-S/16 backbone produced $384$-dimensional features. The slice fusion Transformer used eight attention heads, and the shared MLP classifier used a hidden dimension of $384$ with a dropout rate of $0.25$. We used AdamW with a learning rate of $1.5\times10^{-5}$, weight decay of $0.01$, and a batch size of $8$. The learning rate followed a cosine-decay schedule with $10$ warm-up epochs and a total schedule length of $120$ epochs. Training used mixed precision, with random flipping, rotation, noise perturbation, and Mixup ($\alpha=0.1$, probability $=0.5$) applied only to the training set. Hyperparameters, checkpoints, and ordinal calibration thresholds were selected using the validation set, with early stopping after $35$ epochs without improvement. Final probabilities were obtained through validation-weighted checkpoint ensembling and test-time augmentation. All models were evaluated using identical preprocessing and evaluation protocols.

\subsubsection{Baseline models}

We compared PRISM-Net with representative breast MRI diagnostic models and bilateral imaging models, including BreastMRI-FCDD \citep{oviedo2025cancer}, PBPK DAE-CNN \citep{gravina2024}, DisAsymNet \citep{wang2023disasymnet}, BiGAM-Net \citep{haque2026bigamnet}, and the ODELIA Medical Slice Transformer baseline \citep{mullerfranzes2025odelia}. All models were adapted to the same breast-side three-class label space and evaluated on matched test cases.

\subsubsection{Evaluation metrics}

Performance was evaluated at the breast-side level using Macro AUC, Micro AUC, and QWK, which summarized class-balanced discrimination, overall discrimination, and ordinal agreement across no-lesion, benign, and malignant, respectively. All values were expressed on a percentage scale. Results are reported as the point estimate of the final checkpoint ensemble ± the standard deviation obtained from $2,000$ bootstrap resamples. PRISM-Net was compared with each baseline on matched cases using paired bootstrap testing with $10,000$ iterations and a two-sided significance threshold of p < $0.05$. In each bootstrap iteration, resampling was performed at the patient level, with the left and right breast-side predictions from the same patient kept together as one clustered unit. Cohort membership and reference labels were fixed across models, and no test label informed model selection, calibration, or threshold tuning.

\subsection{Performance on the ODELIA public development dataset}

\begin{table*}[t]
\centering
\caption{Performance comparison on the ODELIA public development dataset.}
\label{tab:odelia_public}
\begin{threeparttable}
\scriptsize
\setlength{\tabcolsep}{3.2pt}
\renewcommand{\arraystretch}{1.12}
\newcommand{\ms}[2]{\mbox{$#1 \pm #2$}}
\newcommand{\bms}[2]{\mbox{$\mathbf{#1 \pm #2}$}}

\begin{tabular*}{\textwidth}{@{\extracolsep{\fill}}lcccccc@{}}
\toprule
\multirow{2}{*}{Method}
& \multicolumn{3}{c}{In-Distribution}
& \multicolumn{3}{c}{Out-of-Distribution} \\
\cmidrule(lr){2-4} \cmidrule(lr){5-7}
& \shortstack{Macro\\AUC}
& \shortstack{Micro\\AUC}
& QWK
& \shortstack{Macro\\AUC}
& \shortstack{Micro\\AUC}
& QWK \\
\midrule

BreastMRI-fcdd~ \citep{oviedo2025cancer}
& \ms{59.90}{6.10^{*}}
& \ms{70.20}{9.75^{*}}
& \ms{10.77}{9.75^{*}}
& \ms{54.11}{4.00^{*}}
& \ms{73.01}{5.42^{*}}
& \ms{2.61}{2.83^{*}} \\

PBPK DAE-CNN~ \citep{gravina2024}
& \ms{56.07}{5.84^{*}}
& \ms{68.46}{6.07^{*}}
& \ms{3.78}{3.45^{*}}
& \ms{50.36}{3.69^{*}}
& \ms{65.80}{2.39^{*}}
& \ms{2.24}{4.01^{*}} \\

DisAsymNet~ \citep{wang2023disasymnet}
& \ms{69.36}{6.34^{*}}
& \ms{81.14}{2.64^{*}}
& \ms{24.02}{20.85^{*}}
& \ms{53.71}{7.71}
& \ms{75.65}{2.74}
& \ms{12.78}{13.17^{*}} \\

BiGAM-Net~ \citep{haque2026bigamnet}
& \ms{62.32}{5.85^{*}}
& \ms{72.43}{6.87^{*}}
& \ms{10.86}{12.13^{*}}
& \ms{50.00}{3.02^{*}}
& \ms{70.87}{4.77^{*}}
& \ms{2.96}{3.65^{*}} \\

ODELIA~ \citep{mullerfranzes2025odelia}
& \ms{76.92}{0.48^{*}}
& \ms{85.56}{0.91^{*}}
& \ms{46.75}{1.62^{*}}
& \ms{60.31}{1.07}
& \ms{77.44}{2.17}
& \ms{29.46}{4.58^{*}} \\

\textbf{Ours}
& \bms{84.11}{2.33}
& \bms{90.64}{1.61}
& \bms{60.94}{5.64}
& \bms{68.51}{4.54}
& \bms{80.74}{2.68}
& \bms{43.45}{7.10} \\

\bottomrule
\end{tabular*}

\begin{tablenotes}
\footnotesize
\item Values are reported as mean $\pm$ SD. AUCs are reported as percentages, and QWK values are scaled by $100$. \textsuperscript{*}$p<0.05$ versus PRISM-Net.
\end{tablenotes}
\end{threeparttable}
\end{table*}

\begin{figure*}[t]
    \centering

    \begin{minipage}{\textwidth}
        \centering

        \begin{subfigure}[t]{0.40\linewidth}
            \vspace{0pt}
            \centering
            \includegraphics[
                height=5.2cm,
                keepaspectratio
            ]{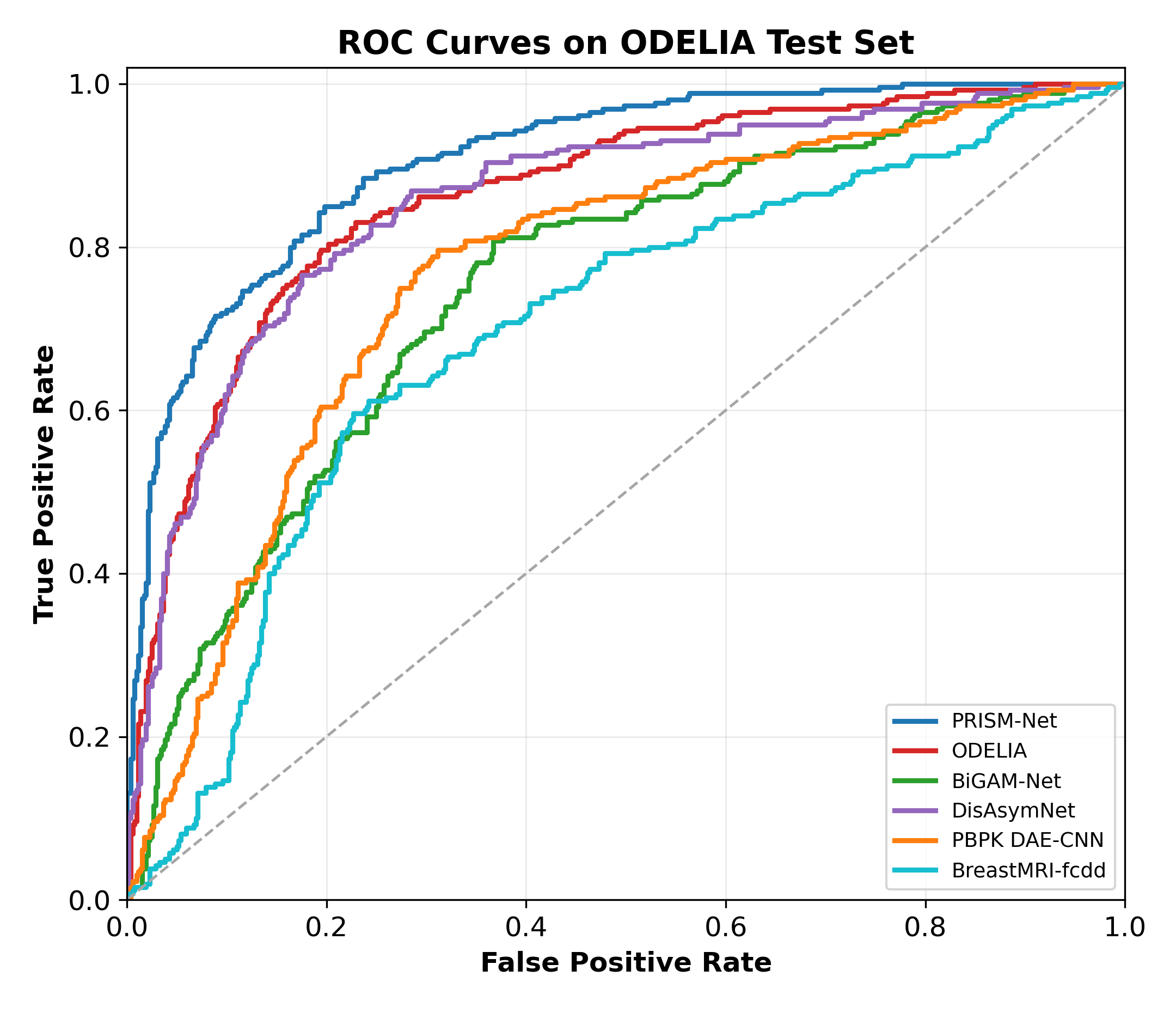}
            \caption{ROC curves.}
            \label{fig:odelia_id_roc}
        \end{subfigure}
        \hspace{0.035\linewidth}
        \begin{subfigure}[t]{0.54\linewidth}
            \vspace{0pt}
            \centering
            \includegraphics[
                height=5.2cm,
                keepaspectratio
            ]{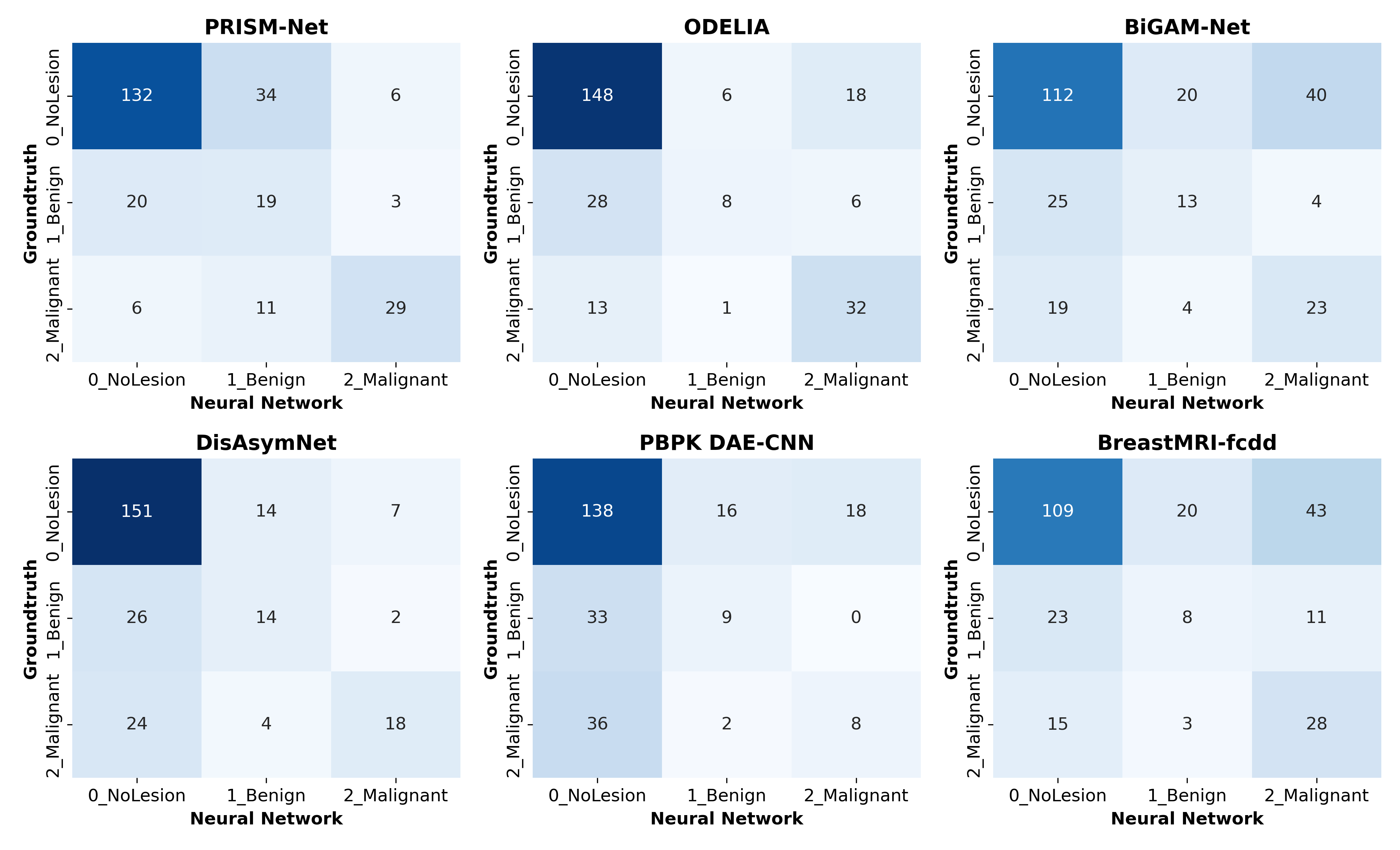}
            \caption{Confusion matrices.}
            \label{fig:odelia_id_confusion}
        \end{subfigure}

    \end{minipage}

    \caption{\textbf{ROC curves and confusion matrices of six models on the ODELIA in-distribution (ID) test set.} The left panel shows the ROC curves, while the right panel presents the corresponding confusion matrices for three-class classification of no-lesion, benign, and malignant.}
    \label{fig:odelia_id_performance}
\end{figure*}

\begin{figure*}[t]
    \centering

    \begin{minipage}{\textwidth}
        \centering

        \begin{subfigure}[t]{0.40\linewidth}
            \vspace{0pt}
            \centering
            \includegraphics[
                height=5.2cm,
                keepaspectratio
            ]{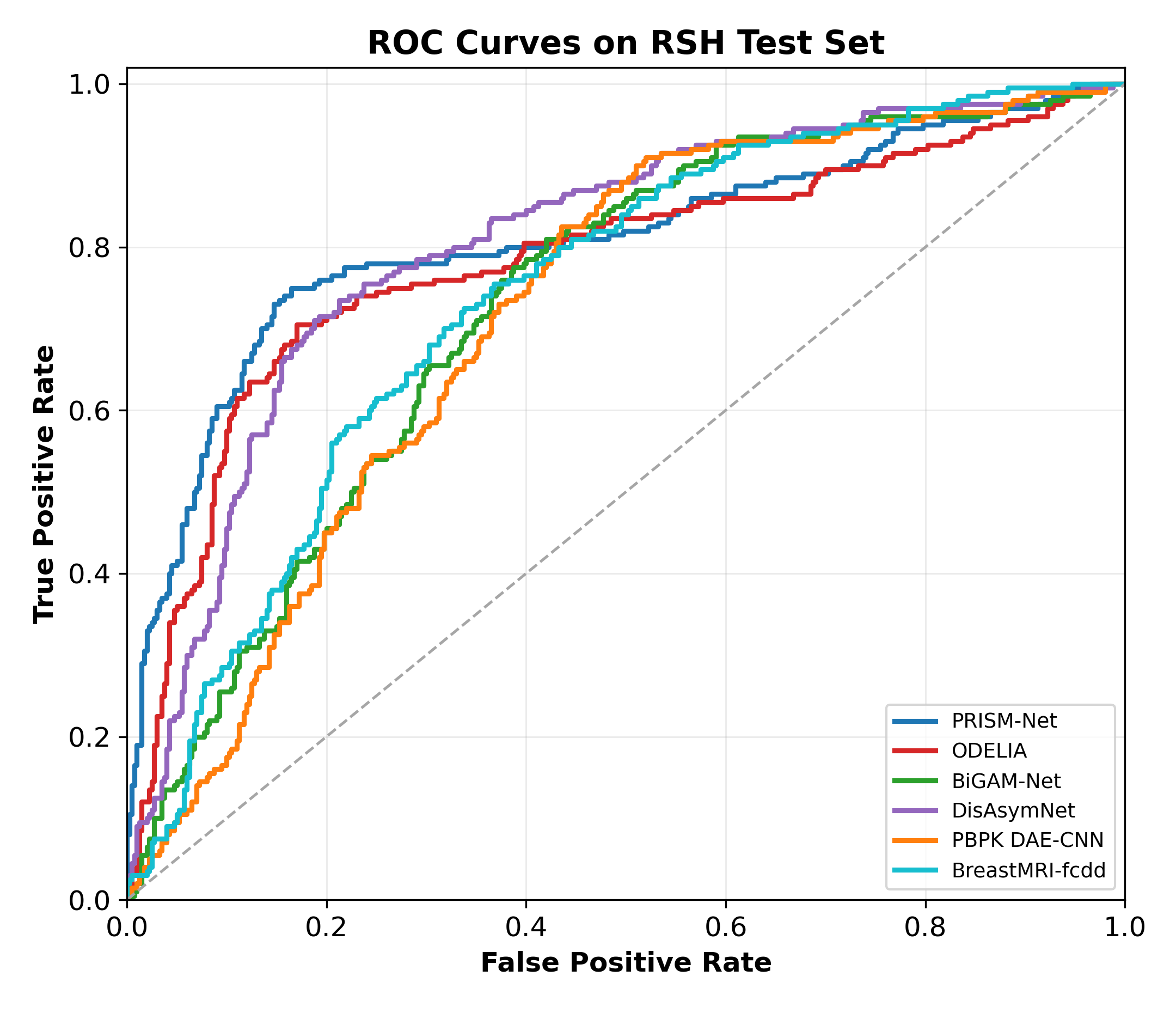}
            \caption{ROC curves.}
            \label{fig:rsh_ood_roc}
        \end{subfigure}
        \hspace{0.035\linewidth}
        \begin{subfigure}[t]{0.54\linewidth}
            \vspace{0pt}
            \centering
            \includegraphics[
                height=5.2cm,
                keepaspectratio
            ]{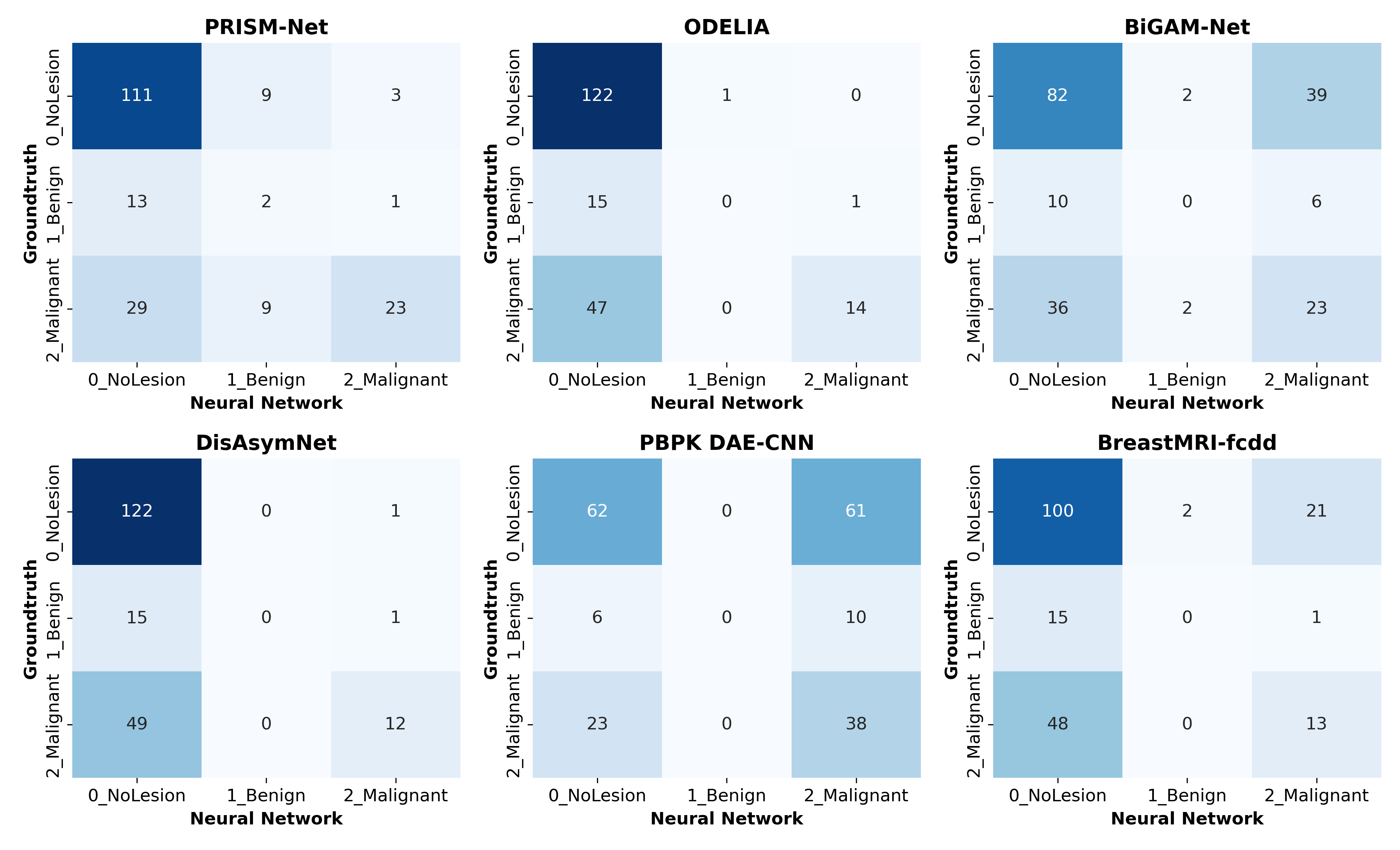}
            \caption{Confusion matrices.}
            \label{fig:rsh_ood_confusion}
        \end{subfigure}

    \end{minipage}

    \caption{\textbf{ROC curves and confusion matrices of six models on the RSH out-of-distribution (OOD) test set.} The left panel shows the ROC curves, while the right panel presents the corresponding confusion matrices for three-class classification of no-lesion, benign, and malignant.}
    \label{fig:rsh_ood_performance}
\end{figure*}

PRISM-Net achieved the best performance on both the ODELIA in-distribution and RSH out-of-distribution sets (Table~\ref{tab:odelia_public}). Its Macro AUC, Micro AUC, and QWK were $84.11 \pm 2.33$, $90.64 \pm 1.61$, and $60.94 \pm 5.64$ in-distribution, and $68.51 \pm 4.54$, $80.74 \pm 2.68$, and $43.45 \pm 7.10$ out-of-distribution. Relative to the strongest baseline, ODELIA, the corresponding gains were $7.19$, $5.08$, and $14.19$ points in-distribution and $8.20$, $3.30$, and $13.99$ points out-of-distribution. ROC curves and confusion matrices are shown in Figs.~\ref{fig:odelia_id_performance} and~\ref{fig:rsh_ood_performance}.

\subsection{External validation on the private institutional test set}

We compared PRISM-Net with DisAsymNet and ODELIA, which were the two strongest baseline methods on the ODELIA public development dataset. The independent 2025 institutional cohort was not used for training, validation, model selection, or tuning. PRISM-Net achieved the highest Macro AUC, Micro AUC, and QWK: $66.85 \pm 3.47$, $86.38 \pm 1.84$, and $38.85 \pm 6.43$, respectively (Table~\ref{tab:private_background_complexity}). Its QWK exceeded ODELIA and DisAsymNet by $40.86$ and $26.41$ points, respectively. The corresponding ROC curves and confusion matrices are shown in panels (a) and (d) of Supplementary Fig.~S1.

\subsection{Performance on the background-complexity external test sets}
The comparison was also restricted to DisAsymNet and ODELIA, the two best-performing baselines in the public-dataset evaluation. PRISM-Net also ranked first on both complexity cohorts (Table~\ref{tab:private_background_complexity}). On the BPE-complexity set, Macro AUC, Micro AUC, and QWK were $66.23 \pm 4.90$, $82.66 \pm 2.91$, and $31.48 \pm 9.36$; on the FGT-complexity set, they were $67.71 \pm 3.38$, $80.11 \pm 2.37$, and $40.88 \pm 6.83$. Corresponding ROC curves and confusion matrices are shown in panels (b,e) and (c,f) of Supplementary Fig.~S1.

\begin{table*}[t]
\centering
\caption{External validation performance on the private institutional test set and background-complexity test sets.}
\label{tab:private_background_complexity}
\begin{threeparttable}
\small
\begin{tabularx}{\textwidth}{@{}ll *{3}{>{\centering\arraybackslash}X}@{}}
\toprule
Cohort & Method & Macro AUC & Micro AUC & QWK \\
\midrule

\multirow{3}{*}{2025 private external test set}
& ODELIA~ \citep{mullerfranzes2025odelia}
& $62.66 \pm 3.63$
& $81.64 \pm 1.95^*$
& $-2.01 \pm 1.12^*$ \\

& DisAsymNet~ \citep{wang2023disasymnet}
& $63.14 \pm 4.21$
& $82.62 \pm 2.18^*$
& $12.44 \pm 5.47^*$ \\

& \textbf{PRISM-Net}
& $\mathbf{66.85 \pm 3.47}$
& $\mathbf{86.38 \pm 1.84}$
& $\mathbf{38.85 \pm 6.43}$ \\

\midrule

\multirow{3}{*}{BPE-complexity test set}
& ODELIA~ \citep{mullerfranzes2025odelia}
& $64.87 \pm 5.63$
& $78.55 \pm 3.28$
& $1.91 \pm 2.19^*$ \\

& DisAsymNet~ \citep{wang2023disasymnet}
& $52.69 \pm 6.15$
& $75.90 \pm 3.75^*$
& $11.75 \pm 7.55^*$ \\

& \textbf{PRISM-Net}
& $\mathbf{66.23 \pm 4.90}$
& $\mathbf{82.66 \pm 2.91}$
& $\mathbf{31.48 \pm 9.36}$ \\

\midrule

\multirow{3}{*}{FGT-complexity test set}
& ODELIA~ \citep{mullerfranzes2025odelia}
& $62.23 \pm 3.30$
& $73.10 \pm 2.26^*$
& $-0.58 \pm 1.07^*$ \\

& DisAsymNet~ \citep{wang2023disasymnet}
& $64.64 \pm 4.16$
& $75.53 \pm 2.48^*$
& $15.21 \pm 5.55^*$ \\

& \textbf{PRISM-Net}
& $\mathbf{67.71 \pm 3.38}$
& $\mathbf{80.11 \pm 2.37}$
& $\mathbf{40.88 \pm 6.83}$ \\

\bottomrule
\end{tabularx}
\begin{tablenotes}
\footnotesize
\item Values are mean $\pm$ SD. AUCs are percentages; QWK values are scaled by $100$. \textsuperscript{*}$p<0.05$ PRISM-Net. BPE, background parenchymal enhancement; FGT, fibroglandular tissue.
\end{tablenotes}
\end{threeparttable}
\end{table*}

\subsection{Ablation studies}

The full PRISM-Net was best in both settings (Table~\ref{tab:ablation_components}). Removing the bilateral asymmetry branch reduced QWK by $14.56$ points in-distribution and $13.40$ points out-of-distribution; removing focal score reweighting reduced QWK by $5.55$ and $14.38$ points, respectively. Macro and Micro AUC also decreased for both variants, supporting the contribution of each component.

\begin{table*}[t]
\centering
\caption{Ablation study of the main components in PRISM-Net on the ODELIA public development dataset.}
\label{tab:ablation_components}
\begin{threeparttable}
\small
\newcommand{\ms}[2]{\mbox{$#1 \pm #2$}}
\newcommand{\bms}[2]{\mbox{$\mathbf{#1 \pm #2}$}}
\begin{tabularx}{\textwidth}{@{}l *{6}{>{\centering\arraybackslash}X}@{}}
\toprule
\multirow{2}{*}{Variant}
& \multicolumn{3}{c}{In-Distribution}
& \multicolumn{3}{c}{Out-of-Distribution} \\
\cmidrule(lr){2-4} \cmidrule(lr){5-7}
& Macro AUC & Micro AUC & QWK
& Macro AUC & Micro AUC & QWK \\
\midrule

PRISM-Net w/o bilateral asymmetry branch
& \ms{79.75}{1.37^*}
& \ms{87.39}{1.34^*}
& \ms{46.38}{5.17^*}
& \ms{64.15}{3.92^*}
& \ms{77.88}{3.53^*}
& \ms{30.05}{3.88^*} \\

PRISM-Net w/o focal score reweighting
& \ms{79.68}{3.22^*}
& \ms{88.12}{2.45^*}
& \ms{55.39}{6.39}
& \ms{64.53}{6.83^*}
& \ms{79.07}{4.58}
& \ms{29.07}{11.53^*} \\

\textbf{PRISM-Net}
& \bms{84.11}{2.33}
& \bms{90.64}{1.61}
& \bms{60.94}{5.64}
& \bms{68.51}{4.54}
& \bms{80.74}{2.68}
& \bms{43.45}{7.10} \\

\bottomrule
\end{tabularx}
\begin{tablenotes}
\footnotesize
\item  Values are mean $\pm$ SD. AUCs are percentages; QWK values are scaled by $100$. \textsuperscript{*}$p<0.05$ versus PRISM-Net. 
\end{tablenotes}
\end{threeparttable}
\end{table*}

Second, we compared different feature extraction backbones while keeping the remaining bilateral learning framework unchanged. With the bilateral framework and evaluation protocol fixed, DINOv3 outperformed ResNet-18, ResNet-50, ViT, and DenseNet-121 on all six metrics (Table~\ref{tab:ablation_backbone}). DenseNet-121 was the second-best backbone. Relative to DenseNet-121, DINOv3 improved Macro AUC, Micro AUC, and QWK by $2.91$, $1.48$, and $6.74$ points in-distribution and by $2.17$, $0.77$, and $3.29$ points out-of-distribution.

\begin{table*}[t]
\centering
\caption{Ablation study of different feature extraction backbones on the ODELIA public development dataset.}
\label{tab:ablation_backbone}
\begin{threeparttable}
\small
\begin{tabularx}{\textwidth}{@{}l *{6}{>{\centering\arraybackslash}X}@{}}
\toprule
\multirow{2}{*}{Backbone}
& \multicolumn{3}{c}{In-Distribution}
& \multicolumn{3}{c}{Out-of-Distribution} \\
\cmidrule(lr){2-4} \cmidrule(lr){5-7}
& Macro AUC & Micro AUC & QWK
& Macro AUC & Micro AUC & QWK \\
\midrule

ResNet-18~ \citep{he2016resnet}
& $80.40 \pm 1.23$
& $88.65 \pm 1.35^*$
& $51.29 \pm 3.49^*$
& $65.49 \pm 4.32$
& $80.34 \pm 2.88$
& $35.84 \pm 5.88^*$ \\

ResNet-50~ \citep{he2016resnet}
& $77.95 \pm 2.90^*$
& $86.89 \pm 0.81^*$
& $48.52 \pm 6.86^*$
& $64.80 \pm 3.17$
& $78.95 \pm 3.54$
& $32.25 \pm 6.35^*$ \\

ViT~ \citep{dosovitskiy2021vit}
& $78.15 \pm 2.06^*$
& $87.27 \pm 2.23^*$
& $50.84 \pm 6.51$
& $61.49 \pm 4.03^*$
& $78.61 \pm 3.01$
& $24.27 \pm 8.58^*$ \\

DenseNet-121~ \citep{huang2017densenet}
& $81.20 \pm 1.67$
& $89.16 \pm 1.46$
& $54.20 \pm 1.48$
& $66.34 \pm 4.51$
& $79.97 \pm 2.47$
& $40.16 \pm 6.79$ \\

\textbf{DINOv3}~ \citep{simeoni2025dinov3}
& $\mathbf{84.11 \pm 2.33}$
& $\mathbf{90.64 \pm 1.61}$
& $\mathbf{60.94 \pm 5.64}$
& $\mathbf{68.51 \pm 4.54}$
& $\mathbf{80.74 \pm 2.68}$
& $\mathbf{43.45 \pm 7.10}$ \\

\bottomrule
\end{tabularx}
\begin{tablenotes}
\footnotesize
\item Values are mean $\pm$ SD. AUCs are percentages; QWK values are scaled by $100$. \textsuperscript{*}$p<0.05$ versus PRISM-Net. 
\end{tablenotes}
\end{threeparttable}
\end{table*}

\subsection{Interpretability and failure cases}

\begin{figure*}[t]
    \centering
    \includegraphics[
        width=0.95\textwidth,
    ]{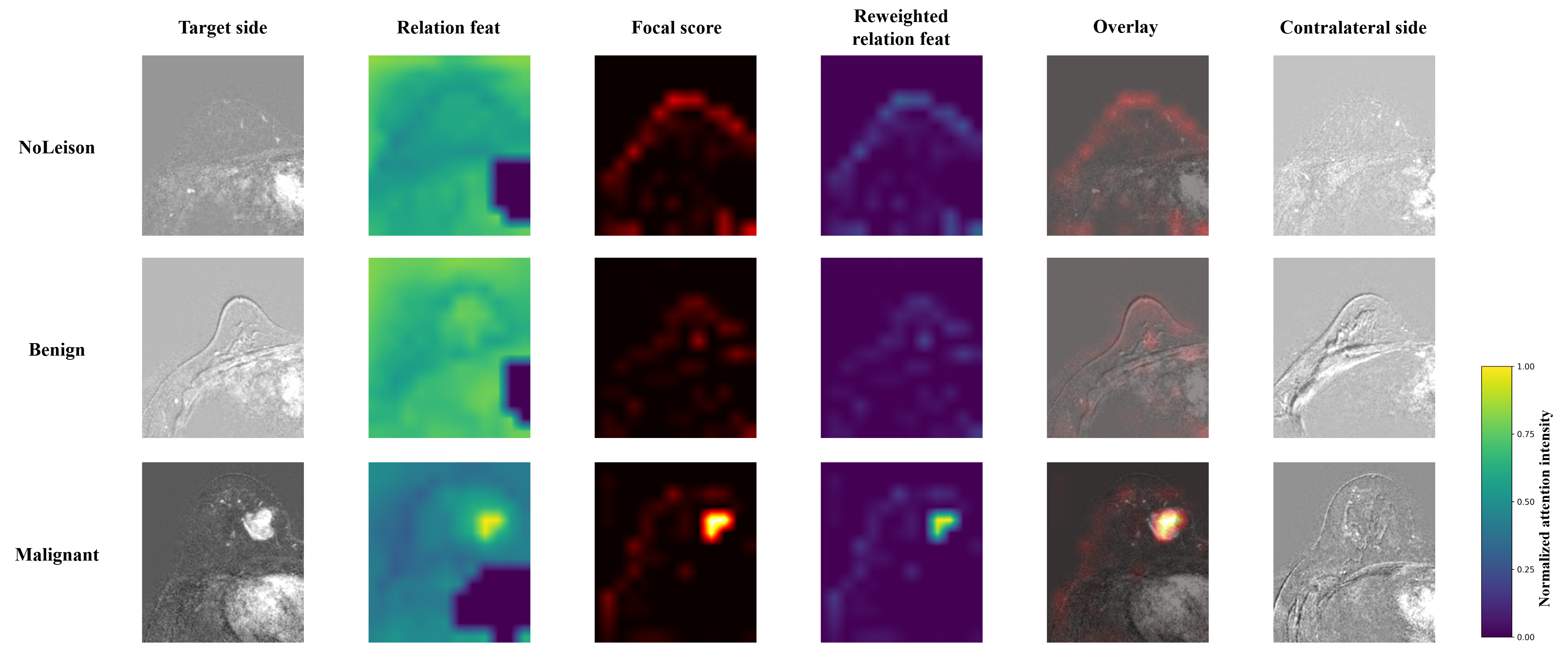}
    \caption{\textbf{Representative interpretability maps generated by PRISM-Net for no-lesion, benign, and malignant cases.} Each row shows one representative case, and each column corresponds to a different stage of the bilateral asymmetry modeling process. From left to right, the columns show the target-side DCE-MRI slice, the relation feature before focal reweighting, the focal score map, the reweighted relation feature, the overlay of the reweighted relation feature on the target-side image, and the contralateral reference side. In the no-lesion case, the focal response is relatively diffuse and does not form a clear lesion-centered activation. In the benign case, the model produces mild and localized asymmetric responses. In the malignant case, the focal score and reweighted relation feature highlight a compact high-response region that spatially corresponds to the suspicious enhancing lesion. These visualizations indicate that PRISM-Net can suppress non-specific bilateral differences and emphasize diagnostically meaningful focal asymmetry.}
    \label{fig:interpretability}
\end{figure*}

Representative maps (Fig.~\ref{fig:interpretability}) show that focal reweighting suppresses diffuse or background-related bilateral differences and concentrates responses on localized asymmetry. The no-lesion case showed diffuse activation, the benign case a mild localized response, and the malignant case a compact high-response region corresponding to the enhancing lesion. These maps provide feature-level explanations rather than lesion segmentations.

\begin{figure*}[t]
    \centering
    \includegraphics[
        width=0.95\textwidth,
    ]{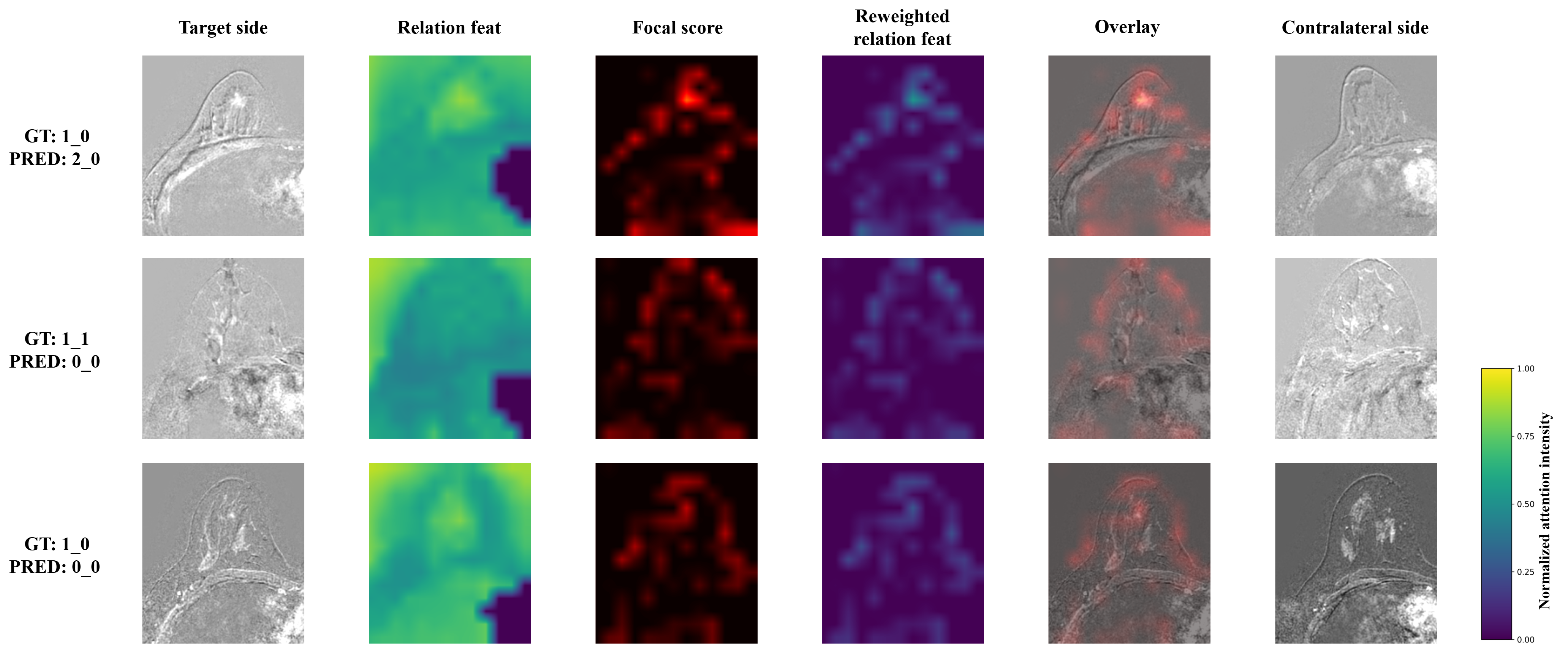}
    \caption{\textbf{Failure-case interpretability and model decision boundaries. }Each row shows one misclassified case. Columns show the target breast, bilateral relation feature, focal score, reweighted relation feature, heatmap, and contralateral breast. GT and PRED denote ground truth and prediction; $0$, $1$, and $2$ represent no-lesion, benign, and malignant, respectively, with suffixes indicating breast side.Row $1$ shows a benign BI-RADS $5$ lesion predicted as malignant. The irregular, spiculated, heterogeneously enhancing lesion produced a strong focal response, while pathology revealed adenosis, sclerosing adenosis, ductal epithelial hyperplasia, and intraductal papilloma, illustrating overlap between benign proliferation and malignancy.Row $2$ shows a benign case in a $39$-year-old woman that was predicted as no-lesion. The patient had multiple bilateral fibroadenomas, dense FGT, and minimal BPE. Enhancing masses were present in both breasts, and corresponding enhancement was also observed in the contralateral side, which weakened the discriminative value of bilateral asymmetry for lesion detection. The Row $3$ shows an approximately $8$-mm fibroadenoma predicted as no-lesion; its small size, circumscribed margin, and dense FGT corresponded to weak focal and reweighted relation responses. These cases highlight three decision boundaries: malignant-like benign proliferation, bilateral lesions with limited asymmetry, and small lesions obscured by complex background signal. }
    \label{fig:failure_cases}
\end{figure*}

We further analyzed representative failure cases to characterize the limitations and decision boundaries of PRISM-Net. As shown in Fig.~\ref{fig:failure_cases}, the errors mainly occurred in three challenging scenarios: benign lesions with malignant-like enhancement patterns, bilateral benign lesions that weakened the discriminative value of contralateral symmetry, and small low-suspicion lesions embedded within dense fibroglandular tissue. These cases suggest that the model may still be affected by overlapping enhancement appearances between benign and malignant lesions, confusing bilateral enhancement patterns, and limited lesion conspicuity in complex backgrounds.

%% =============================================================================
\section{Discussion}
\label{sec:discussion}
This study developed PRISM-Net as a registration-free bilateral learning framework for three-class breast DCE-MRI classification. The main finding is that explicitly modeling the contralateral breast as a patient-specific bilateral reference representation improved both discriminative performance and ordinal diagnostic agreement across public, external, and background-complexity test settings. On the ODELIA in-distribution test set, PRISM-Net achieved a Macro AUC of $84.11 \pm 2.33$, Micro AUC of $90.64 \pm 1.61$, and QWK of $60.94 \pm 5.64$. This advantage was preserved on the RSH out-of-distribution institution. These results suggest that PRISM-Net did not merely improve class separability, but also better preserved the ordered relationship among no-lesion, benign, and malignant. 

The observed gains may reflect the way PRISM-Net uses bilateral information for patient-specific background normalization rather than simply treating the contralateral breast as an additional input. Radiological bilateral comparison benefits from shared patient-level physiological and acquisition conditions, but the contralateral breast cannot be assumed to be anatomically identical or disease-free, avoiding the assumption of anatomical symmetry or disease-free status. PRISM-Net therefore treats it as an approximate comparative context rather than a normal template. Shared-weight encoding maps both breasts into a comparable feature space, body-mask-guided patch filtering restricts matching to anatomically plausible breast tissue, and adaptive patch-level matching reduces dependence on voxel-wise registration despite differences in breast shape, positioning, and deformation. The interpretability analysis in Fig.~\ref{fig:interpretability} provides qualitative support for this mechanism. Before focal reweighting, the bilateral relation features contained relatively diffuse responses that could reflect both lesion-related asymmetry and nonspecific background variation. After reweighting, the responses became more spatially concentrated: the malignant example showed a compact high-response region spatially overlapping with the suspicious enhancing region, the no-lesion example lacked a distinct lesion-centered activation, and the benign example exhibited an intermediate pattern. These observations suggest that the bilateral focal response module may help reduce sensitivity to diffuse or bilaterally shared variations while emphasizing localized asymmetric features that are more relevant to classification, which is consistent with the model design. Nevertheless, these visualizations should be interpreted as qualitative feature-level explanations. 

The background-complexity analyses further indicate that bilateral comparison may be particularly useful when unilateral enhancement patterns are confounded by complex parenchymal backgrounds. Moderate or marked BPE has previously been associated with reduced diagnostic performance on breast MRI \citep{bechynaBaltzer2025}. In the BPE- and FGT-complexity test sets, PRISM-Net consistently achieved the highest performance among evaluated methods, with the most pronounced relative gains observed for QWK, as shown in Table~\ref{tab:private_background_complexity}. The confusion matrices in Supplementary Fig. S1 further illustrate fewer severe ordinal misclassifications. As shown in Fig. 6 benign case, dense FGT confounded radiological interpretation in a representative clinical case, leading to a BI-RADS 5 assessment. PRISM-Net, however, correctly classified the affected breast as benign, consistent with postoperative histopathology showing sclerosing adenosis with intraductal papilloma. These findings suggest that patient-specific bilateral reference modeling may provide valuable comparative information under complex background conditions. Physiological and hormonal factors, including menstrual cycle variation, pregnancy, and lactation, can contribute to increased BPE and heterogeneous background enhancement, potentially reducing lesion conspicuity and increasing diagnostic uncertainty. By incorporating contralateral breast information as an individualized comparative reference, PRISM-Net better capture clinically relevant asymmetric patterns beyond absolute enhancement characteristics in such challenging examinations. This capability may be relevant for lesions presenting as NME, where the absence of a discrete mass and substantial overlap between malignant, benign, and physiological enhancement patterns often complicate interpretation. Nevertheless, performance variation across these challenging sets may be influenced by limited sample sizes and the inherent uncertainty associated with complex background enhancement patterns. Further study in larger cohorts is required.

The ablation studies clarify not only the contribution of individual modules but also the design logic underlying PRISM-Net. As shown in Table~\ref{tab:ablation_components}, removing the bilateral asymmetry branch consistently degrades performance, indicating that the contralateral breast contributes more than additional image context. Instead, it serves as a patient-specific reference that enables the model to distinguish shared physiological enhancement patterns from localized abnormal deviations. This result supports adaptive reference matching in the feature space as a more appropriate formulation of bilateral comparison than unilateral analysis or simple bilateral feature aggregation. The contribution of focal score reweighting further suggests that bilateral differences are not uniformly informative. Diffuse enhancement variation and background related mismatch can generate nonspecific responses, whereas deviations that are locally salient are more likely to contain discriminative evidence. Focal reweighting therefore acts as an evidence selection mechanism that suppresses diffuse interference and concentrates the relational representation on diagnostically relevant asymmetry. The backbone comparison provides a complementary observation that reliable registration free correspondence depends on the semantic quality of the shared feature space, since more discriminative representations facilitate stable cross breast matching under anatomical, positional, and background variability. Taken together, these findings reveal a coherent computational pathway in PRISM-Net: robust feature encoding establishes a comparable bilateral representation space, adaptive matching constructs patient specific references, and focal reweighting identifies the most informative reference conditioned deviations. The performance gains therefore arise not merely from introducing bilateral input, but from structuring bilateral comparison into representation learning, correspondence estimation, and selective evidence extraction. 

Several limitations should be acknowledged. First, although the study used a public development dataset, an out-of-distribution validation institution, and an independent private external test set, the private external data were collected from a single institution. Prospective multicenter validation is therefore needed to assess generalizability across centers, imaging devices, and acquisition protocols. Second, the registration-free bilateral design should be interpreted as a feature-space approximation of bilateral comparison rather than as explicit anatomical correspondence. Its effectiveness may be reduced when the contralateral breast provides an unreliable reference, which may be compromised by bilateral disease, marked asymmetric background parenchymal enhancement, implants, postoperative changes, motion artifacts, failed fat suppression, or asymmetric image quality. The observed failure cases were consistent with this boundary (Fig.~\ref{fig:failure_cases}). Third,the model was trained at the breast-side level without requiring lesion-level annotations. Although this weakly supervised setting is practical for large-scale clinical data, it limits direct evaluation of lesion localization. It also does not fully exploit complementary MRI information, including T2-weighted imaging, DWI, ADC maps, ultrafast kinetic features, and clinical risk factors.

Future work should extend the present retrospective evaluation through prospective multicenter studies, broader protocol heterogeneity, and reader studies in realistic clinical workflows. Further studies should also evaluate whether incorporating multiparametric MRI sequences, kinetic information, and clinical risk factors can improve diagnostic performance beyond early DCE-MRI alone. In addition, further comparison between asymmetry maps, model attention, and radiologist-annotated lesion locations may help clarify the spatial interpretability. Overall, the present findings support registration-free bilateral comparison as a potential strategy for improving AI-assisted breast DCE-MRI triage. Larger prospective multicenter studies are required to determine its clinical utility.

\section{Conclusion}\label{conclusion}
PRISM-Net presents a registration-free bilateral learning framework for breast-side three-class DCE-MRI classification that treats the contralateral breast as an approximate patient-specific reference rather than a normal template. Through adaptive patch-level matching and focal asymmetry reweighting, the framework leverages bilateral contextual information to characterize diagnostically relevant asymmetry under complex background variation. Across public, external, and background-complexity evaluations, PRISM-Net achieved superior performance compared with existing baselines, while ablation studies demonstrated the complementary contributions of bilateral reference modeling and focal asymmetry reweighting. These findings suggest that patient-specific bilateral comparison may provide a promising strategy for improving AI-assisted interpretation of challenging breast MRI examinations. Further prospective multicenter validation and reader studies are warranted to assess its potential clinical utility.

\section*{Acknowledgements}
\vspace{1.5ex}
\section*{CRediT authorship contribution statement}
% CRediT authorship contribution statement removed for peer review.
Boya Zhang: Conceptualization, Methodology, Investigation, Data curation, Validation, Formal analysis, Funding acquisition, Writing – original draft. Shuaiwen Zhou: Conceptualization, Methodology, Software, Investigation, Formal analysis, Validation, Visualization, Data curation, Writing – original draft. 
Di Kong: Review \& editing, Methodology, Supervision, Conceptualization.
Mingxu Wang: Data curation, Review \& editing.
Wenbiao Du: Data curation, Review \& editing.
Yiman Zhong: Data curation, Review \& editing.
Yuexin Duan: Data curation, Review \& editing.
Xiawei Yue: Data curation, Review \& editing.
Liuquan Cheng: Supervision, Resources, Project administration, Conceptualization, Review \& editing.
Xiru Li: Supervision, Funding acquisition, Resources, Project administration, Review \& editing.

\section*{Ethics statement}
% All procedures and protocols were approved by the Ethics Committee. Ethics statement removed for peer review.
This study involved human subjects and was conducted in accordance with ethical standards. All procedures and protocols were approved by the Ethics Committee of the General Hospital of the People’s Liberation Army (Approval No. S2026-078-01).

\section*{Funding}
% Funding information removed for peer review.
This work is supported by the Zhongguancun Academy (Grant No. 02012411).

\section*{Declaration of competing interest}
The authors declare that they have no known competing financial interests or personal relationships that could have appeared to
influence the work reported in this paper.

\section*{Data and code availability}
The public dataset used for model training and evaluation is accessible through the official ODELIA data.The source code will be made publicly available upon acceptance of this article.

\bibliographystyle{elsarticle-num} 
\bibliography{cas-refs}

% \begin{thebibliography}{9}

% \end{thebibliography}
\end{document}

\let\WriteBookmarks\relax
\def\floatpagepagefraction{1}
\def\textpagefraction{.001}
\appendix

\renewcommand{\thetable}{S\arabic{table}}
\setcounter{table}{0}
\renewcommand{\thefigure}{S\arabic{figure}}
\setcounter{figure}{0}
\FloatBarrier

\begin{table*}[t]
\centering
\caption{DCE-MRI acquisition protocols of the public and private cohorts.}
\label{tab:mri_protocol}
\begin{threeparttable}
\small
\begin{tabularx}{\textwidth}{@{}lXX@{}}
\toprule
Parameter & Public multicenter cohort & Private institutional cohort \\
\midrule
Dataset
& ODELIA public breast MRI dataset
& Private institutional breast DCE-MRI cohort \\

Scanner vendor
& GE, Siemens, and Philips
& GE Discovery MR750 \\

Field strength
& 1.5 T and 3.0 T
& 3.0 T \\

Breast coil
& Dedicated bilateral breast coils, 4--18 channels depending on center
& 8-channel dedicated breast coil \\

Sequence used for model input
& Dynamic T1-weighted sequence
& VIBRANT dynamic T1-weighted imaging \\

Imaging plane
& Axial
& Axial \\

TR/TE, ms
& 4.4--7.1 / 1.7--4.6 for most 3D protocols
& 7.7 / 4.3 \\

Flip angle
& 9.8--18$^\circ$ for most 3D protocols
& 10$^\circ$ \\

Field of view, mm
& 201--427
& 320 \\

Acquisition matrix
& 256--672
& 320 $\times$ 320 \\

Slice thickness, mm
& 1.0--3.1
& 1.0 \\

Post-contrast phases
& 2--7
& 4 \\

Contrast dose
& 0.1 mmol/kg
& 0.1 mmol/kg \\

Injection rate
& 1--3 mL/s
& 2 mL/s \\

Saline flush
& 25--30 mL
& 20 mL \\

\bottomrule
\end{tabularx}
\begin{tablenotes}
\footnotesize
\item Public-cohort values are summarized as ranges across centers from the original dataset supplementary material. DCE-MRI, dynamic contrast-enhanced magnetic resonance imaging; TE, echo time; TR, repetition time.
\end{tablenotes}
\end{threeparttable}
\end{table*}
\begin{figure*}[!t]
    \centering
    \captionsetup[subfigure]{font=small,skip=3pt}

    % =========================================================
    % 第一排：三个 ROC 图
    % =========================================================
    \begin{subfigure}[t]{0.315\textwidth}
        \centering
        \includegraphics[
            width=\linewidth,
            keepaspectratio
        ]{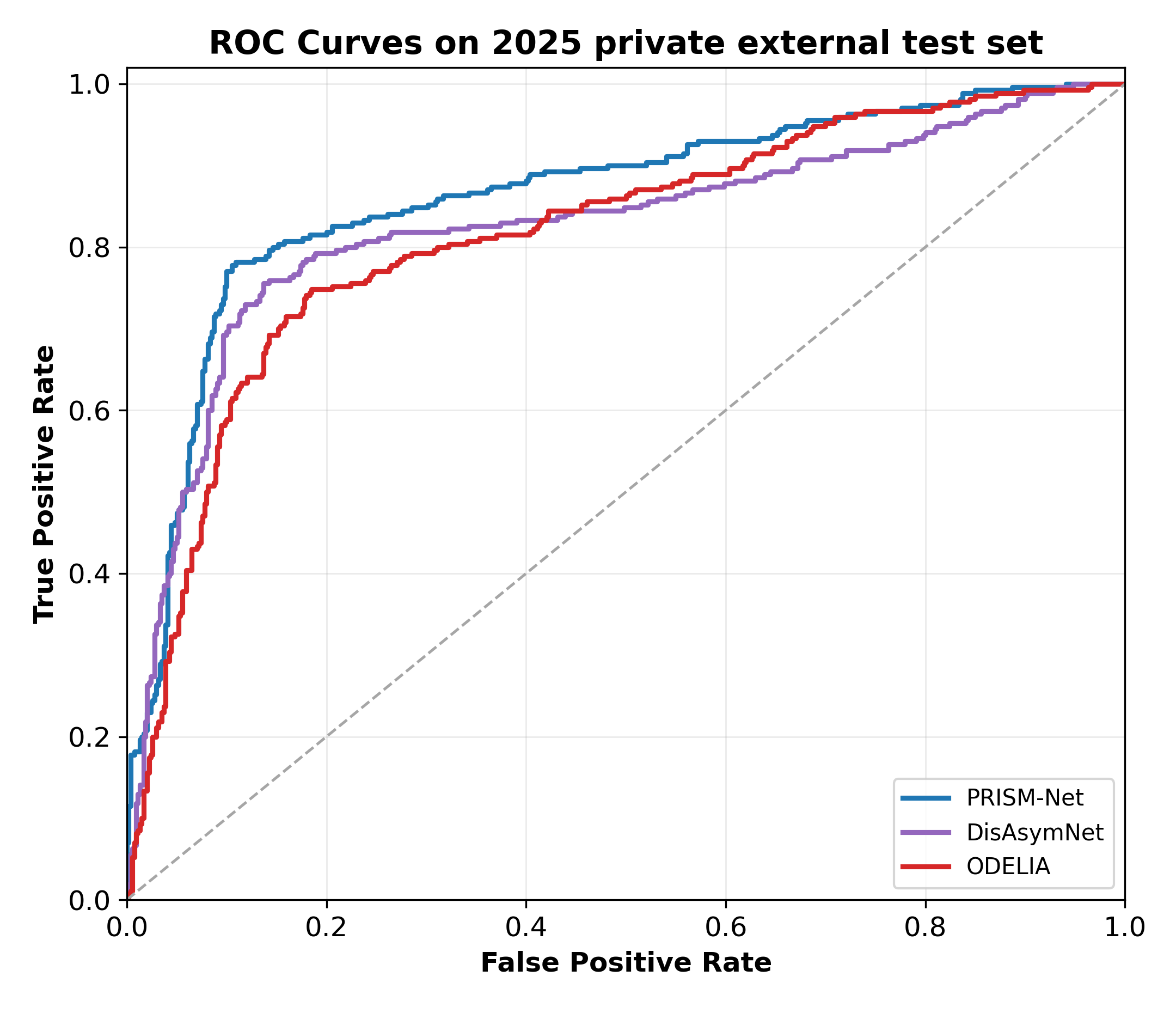}
        \caption{2025 private external test set.}
        \label{fig:private_overall_roc}
    \end{subfigure}
    \hfill
    \begin{subfigure}[t]{0.315\textwidth}
        \centering
        \includegraphics[
            width=\linewidth,
            keepaspectratio
        ]{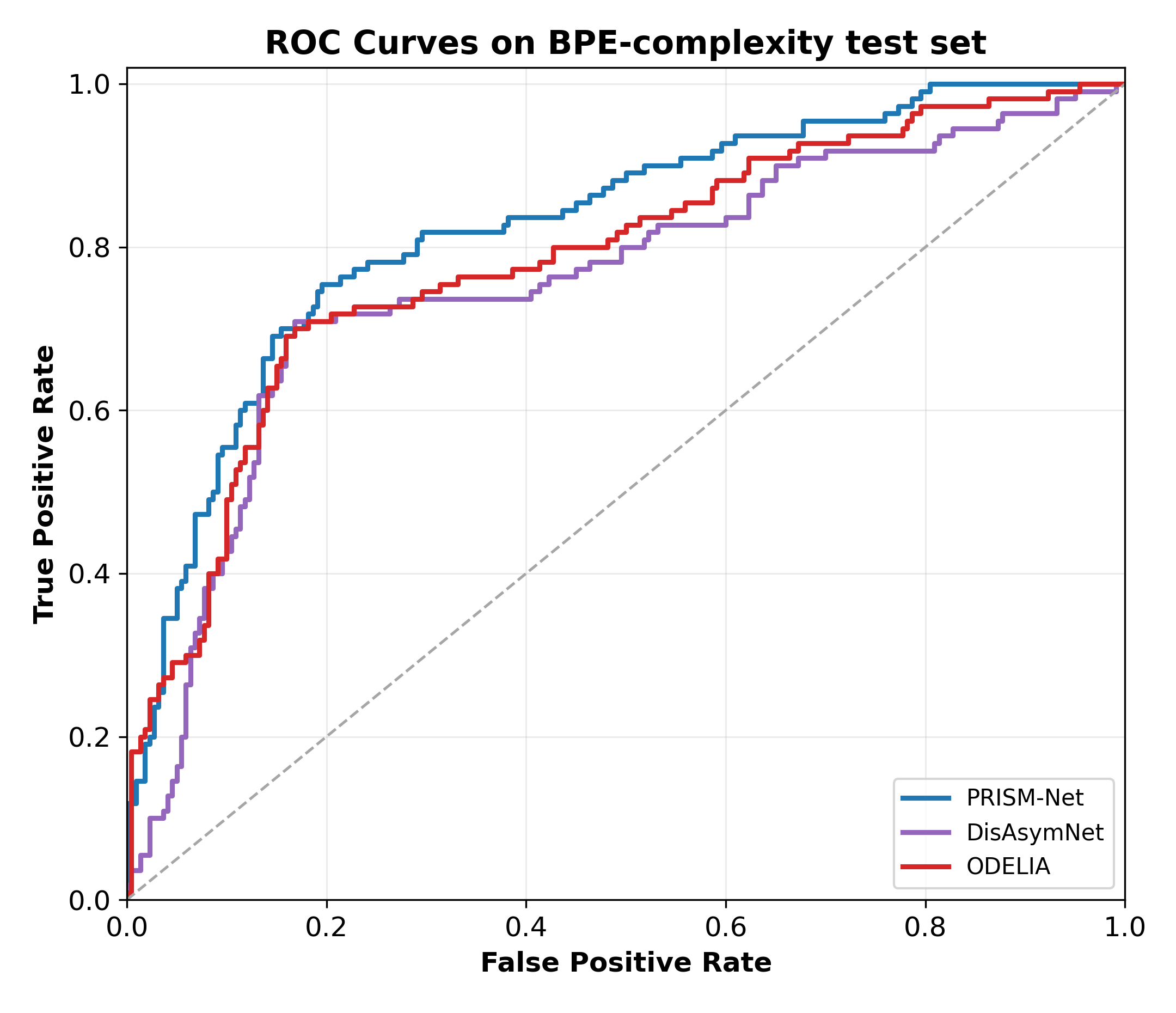}
        \caption{BPE-complexity test set.}
        \label{fig:bpe_difficult_roc}
    \end{subfigure}
    \hfill
    \begin{subfigure}[t]{0.315\textwidth}
        \centering
        \includegraphics[
            width=\linewidth,
            keepaspectratio
        ]{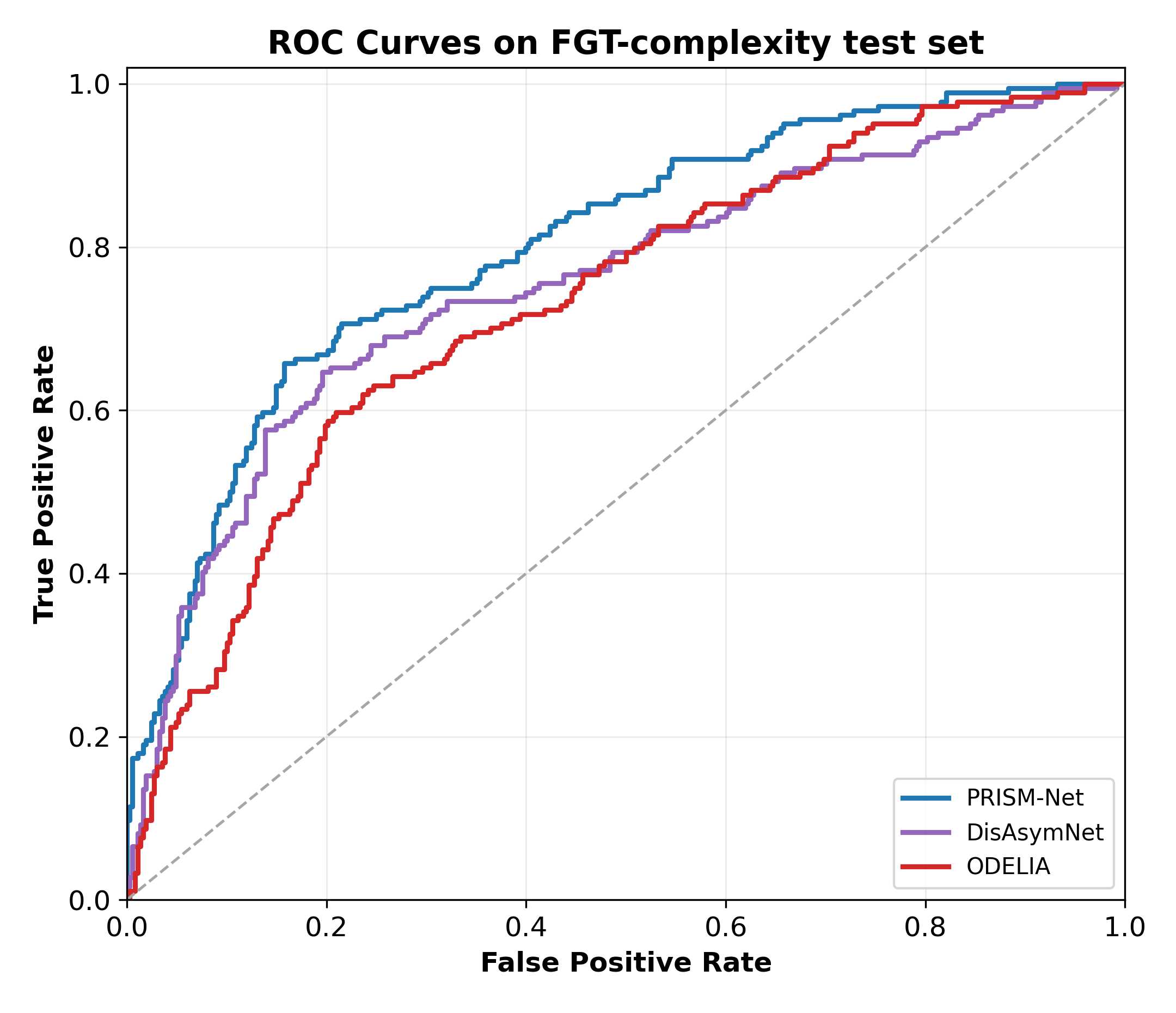}
        \caption{FGT-complexity test set.}
        \label{fig:fgt_difficult_roc}
    \end{subfigure}

    \vspace{0.5em}

    % =========================================================
    % 第二排：总体私有数据集混淆矩阵
    % =========================================================
    \begin{subfigure}[t]{0.98\textwidth}
        \centering
        \includegraphics[
            width=0.98\linewidth,
            keepaspectratio
        ]{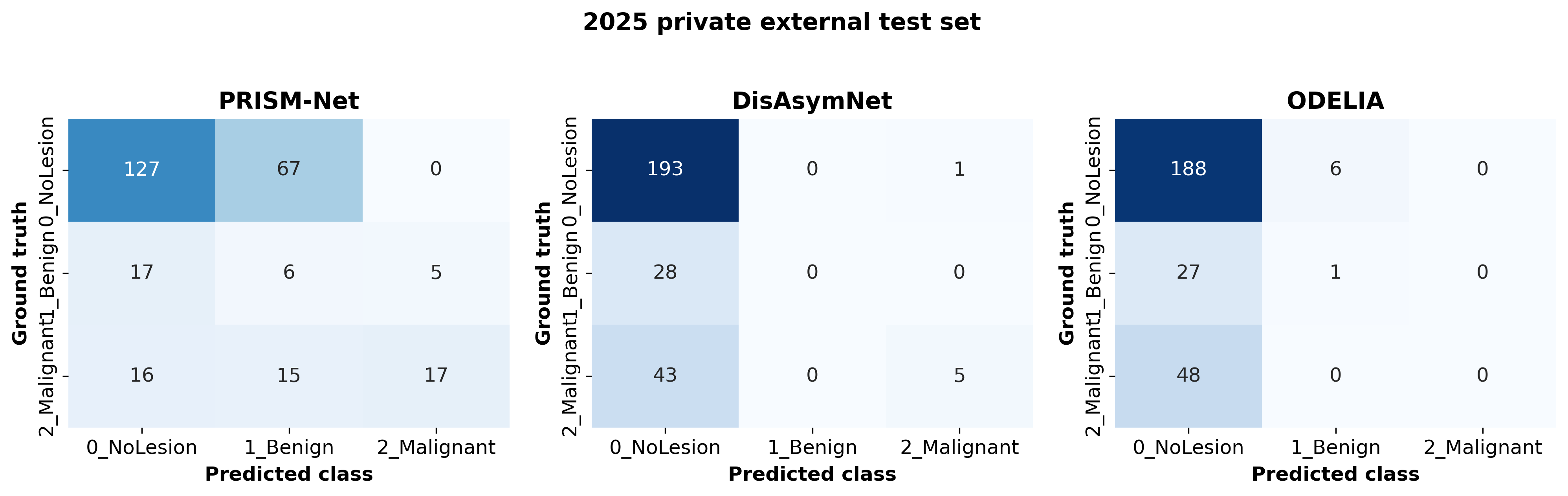}
        \caption{Confusion matrices on the 2025 private external test set.}
        \label{fig:private_overall_confusion}
    \end{subfigure}

    \vspace{0.4em}

    % =========================================================
    % 第三排：BPE 困难亚组混淆矩阵
    % =========================================================
    \begin{subfigure}[t]{0.98\textwidth}
        \centering
        \includegraphics[
            width=0.98\linewidth,
            keepaspectratio
        ]{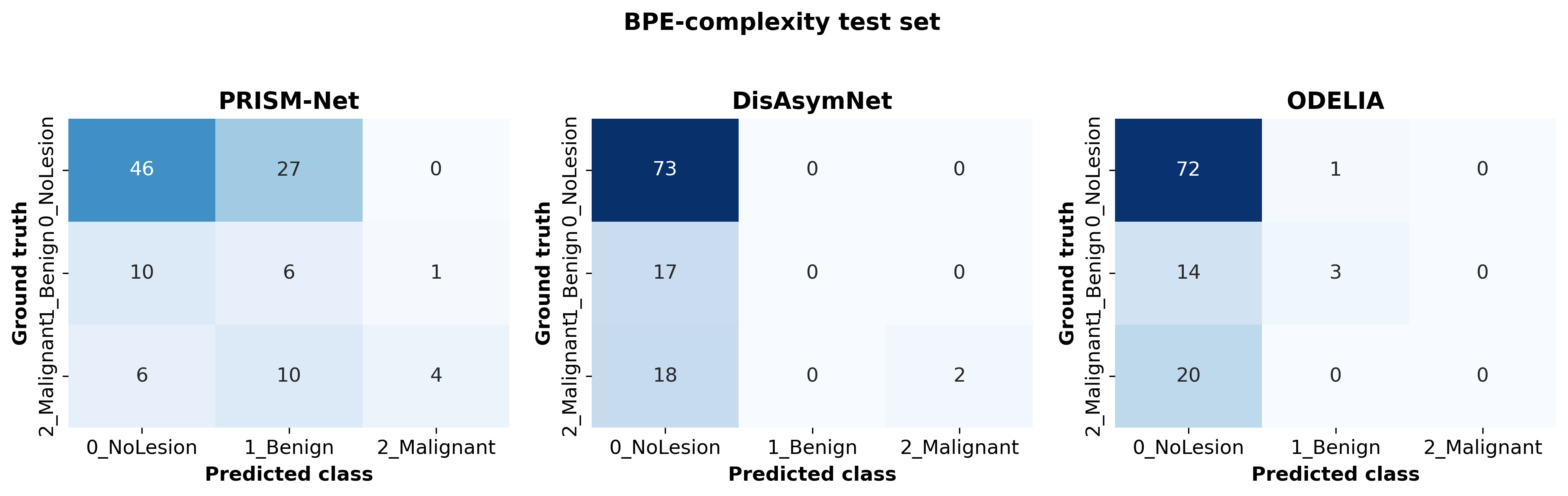}
        \caption{Confusion matrices on the BPE-complexity test set.}
        \label{fig:bpe_difficult_confusion}
    \end{subfigure}

    \vspace{0.4em}

    % =========================================================
    % 第四排：FGT 困难亚组混淆矩阵
    % =========================================================
    \begin{subfigure}[t]{0.98\textwidth}
        \centering
        \includegraphics[
            width=0.98\linewidth,
            keepaspectratio
        ]{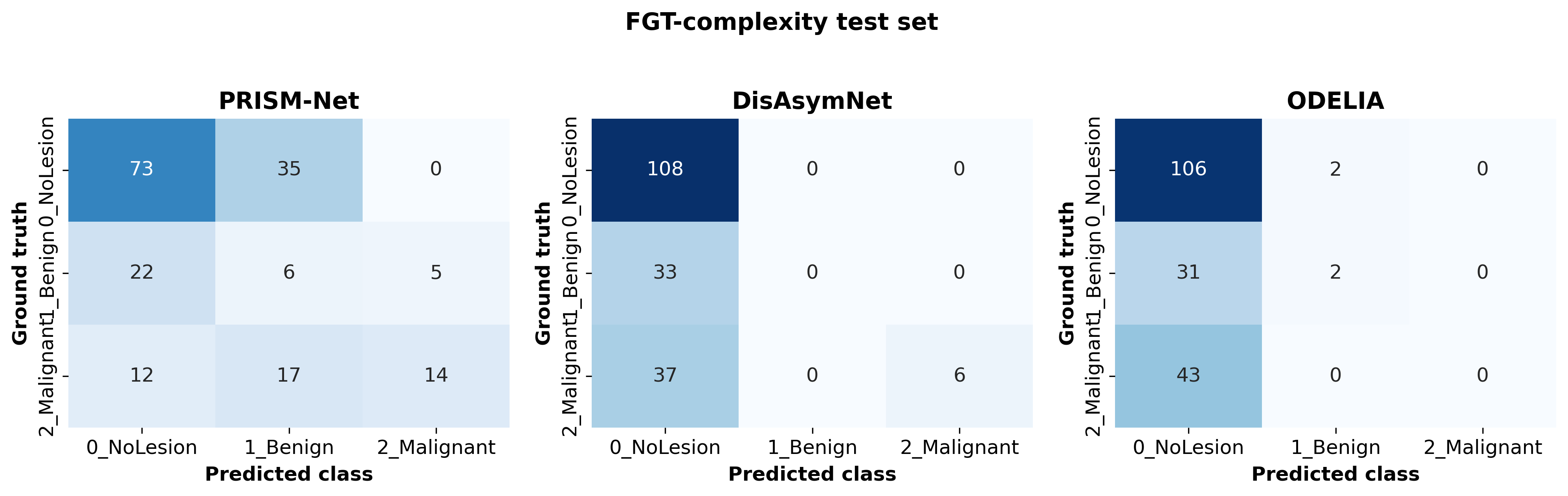}
        \caption{Confusion matrices on the FGT-complexity test set.}
        \label{fig:fgt_difficult_confusion}
    \end{subfigure}

    \caption{
    ROC curves and confusion matrices of PRISM-Net, DisAsymNet, and ODELIA
    on the private institutional cohorts.
    Panels (a)--(c) show the ROC curves on the 2025 private external test set,
    the BPE-complexity test set, and the FGT-complexity test
    set, respectively.
    Panels (d)--(f) present the corresponding confusion matrices for
    three-class classification.
    }
    \label{fig:private_cohort_performance}
\end{figure*}

%% --- Short metadata (for running headers) ---
\shorttitle{PRISM-Net for Breast DCE-MRI}
\shortauthors{Zhang et~al.}